\documentclass[11pt,a4paper]{article}
\usepackage[hyperref]{acl2020}
\usepackage{times}
\usepackage{latexsym}
\usepackage{graphicx}
\usepackage{caption}
\usepackage{subcaption}
\usepackage{amsmath}
\usepackage{amssymb}
\usepackage{multirow}
\DeclareMathOperator*{\argmax}{arg\,max}

\DeclareRobustCommand\onedot{\futurelet\@let@token\@onedot}
\def\onedot{. }
\def\eg{\emph{e.g}\onedot}

\def\etc{\emph{etc.}\onedot}

\usepackage[ruled,linesnumbered]{algorithm2e}

\usepackage{url}

\usepackage{microtype}

\aclfinalcopy 

\usepackage{color}

\newcommand{\emailcustom}[2]{ {\small \href{mailto://#1}{\texttt{#2}}}}

\title{User Memory Reasoning for Conversational Recommendation}

\author{
\AND
Hu Xu\textsuperscript{\text{1\thanks{\hspace{3pt}Most work is done while the first author is a research intern at Facebook.}}}, 
Seungwhan Moon\textsuperscript{\text{2}}, 
Honglei Liu\textsuperscript{\text{2}},
Bing Liu\textsuperscript{\text{2}},
Pararth Shah\textsuperscript{\text{2}}
\AND
Bing Liu\textsuperscript{\text{1,3}}\and Philip S. Yu\textsuperscript{\text{1,4}}\\
    \textsuperscript{1}{Department of Computer Science, University of Illinois at Chicago}\\
    \textsuperscript{2}{Facebook Assistant}\\
    \textsuperscript{3}{WICT, Peking University}\\
    \textsuperscript{4}{Institute for Data Science, Tsinghua University}\\
    \emailcustom{hxu48@uic.edu}{\{hxu48, liub, psyu\}@uic.edu}, \emailcustom{shanemoon@fb.com}{\{shanemoon, honglei, bingl, pararths\}@fb.com}\\
}

\date{}

\begin{document}
\maketitle
\begin{abstract}
We study a conversational recommendation model which dynamically manages users' past (offline) preferences and current (online) requests through a structured and cumulative \textit{user memory knowledge graph}, to allow for natural interactions and accurate recommendations.
For this study, we create a new \textbf{M}emory \textbf{G}raph (MG) $\leftrightarrow$ \textbf{Conv}ersational \textbf{Rec}ommendation parallel corpus called \textit{MGConvRex} with 7K+ human-to-human role-playing dialogs, grounded on a large-scale user memory bootstrapped from real-world user scenarios.
MGConvRex captures human-level reasoning over user memory and has disjoint training/testing sets of users for zero-shot (cold-start) reasoning for recommendation.
We propose a simple yet expandable formulation for constructing and updating the MG, and a reasoning model that predicts optimal dialog policies and recommendation items in unconstrained graph space.
The prediction of our proposed model inherits the graph structure, providing a natural way to explain the model's recommendation.
Experiments are conducted for both offline metrics and online simulation, showing competitive results.
\footnote{The dataset, code and models will be released for future research.}
\end{abstract}

\section{Introduction}
\begin{figure}[t]
\centering
\includegraphics[width=0.9\columnwidth]{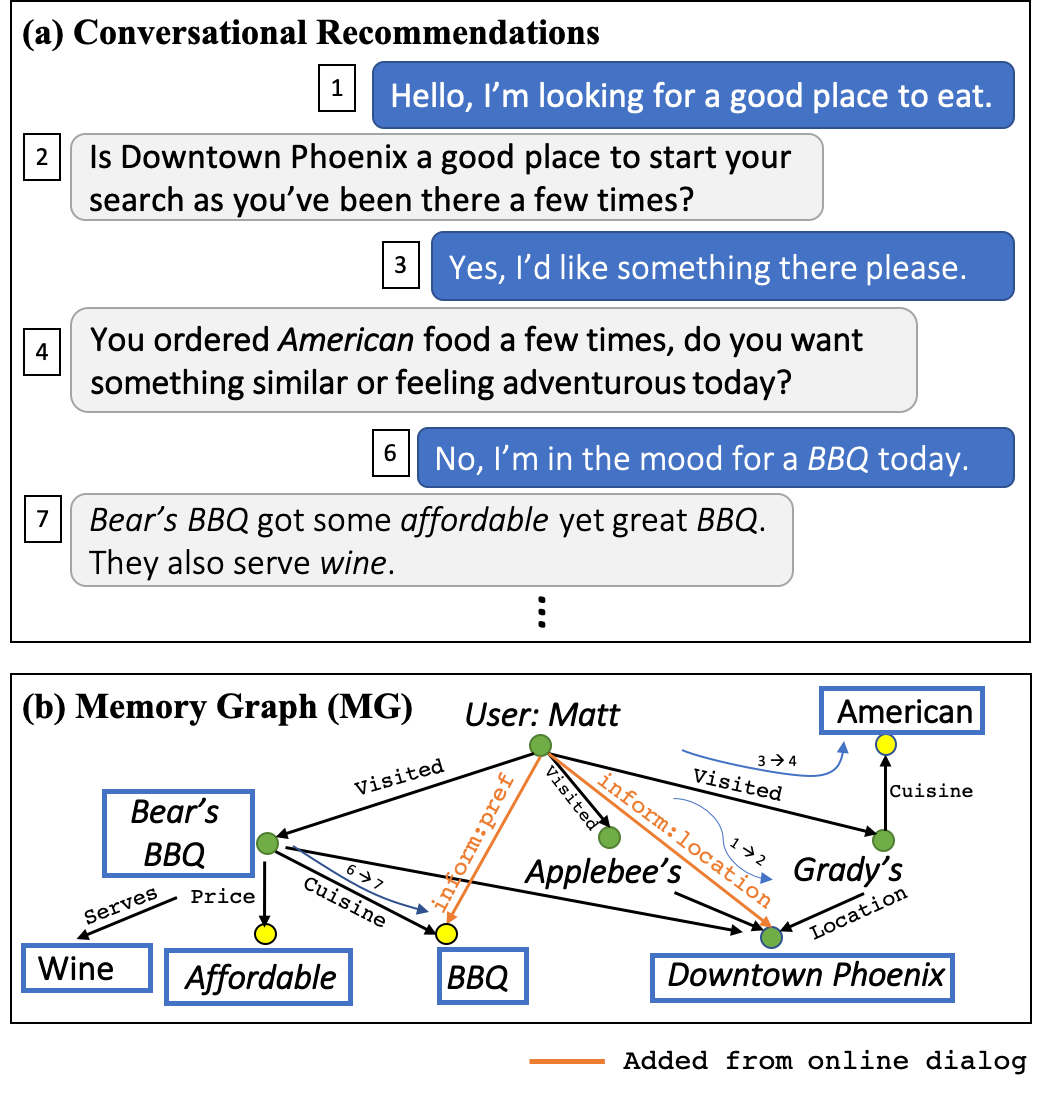}
    \caption{A conceptual illustration of \textbf{Memory-grounded conversational recommendation}. (1) Past (offline) user preferences are captured as an initial Memory Graph (MG). (2) Conversational recommendation allows users to express preferences and requirements through dialogs. (3) Our \textit{MGConvRex} corpus is grounded on user memory, which represents user's past history as well as newly added preferences.}
\label{fig:dialog}
\vspace{-3mm}
\end{figure}

Conversational recommendation system has recently gained traction in the dialog community, in which the model aims to learn up-to-date (online) user preferences, instead of using static (offline) preferences as in the traditional recommender systems (\eg collaborative filtering (CF)).
Most existing works focus on combining a static recommender system with a dialog system by updating user preferences via asking relevant questions (often referred as ``System Ask User Respond (SAUR)'' \cite{zhang2018towards}.
However, this ``short-term'' update in the model unnaturally isolates users' history and their preference in the current dialog (that are possibly forgotten after the dialog is finished).
An intelligent system should be able to dynamically maintain and reason over users' knowledge for current (and possibly future) recommendations.

To this end, we introduce a novel concept called \textit{user memory graph} to holistically represent the knowledge about users and associated items.
This user memory graph may contain any static knowledge obtained offline (\eg items, attributes, the history of users and past dialogs) and users' knowledge online (\eg from state tracking of the current dialog), as illustrated in Figure \ref{fig:dialog}.
User memory graph naturally has the following benefits.
(1) \textit{Holistic reasoning} considers available knowledge about users and items all together to generate dialog policy.
We believe this is the core problem in conversational recommendation
because asking a good question or finding a good candidate item needs to explore the ``soft match'' of the knowledge between users and items\footnote{In contrast, task-oriented dialog has a focus on hard constraints matching (\eg DB query) on available records, although their differences can be blurry.}\cite{zhang2018towards}. 
(2) \textit{Zero-shot (cold-start) reasoning} for users/items unseen during training. 
User memory graph naturally separates user/item knowledge from the reasoning process of policy.
As a result, one can train a user/item agnostic model that can be later applied to the user memory graph for a new user (obtained after the model is deployed).
In contrast, most CF-based system ``overfits'' to existing users / items (in their embeddings).
(3) \textit{Open space policy} is a key challenge in conversational recommendation because of the innumerable items involved in dialog policy.
This requires a flexible space of policy to cover all items (and possibly all valid values and slots
\footnote{We widely reuse the terms from task-oriented dialog to make this paper easier to read, although slots and values can be special cases of entities in a user knowledge graph.} for acquiring preference) instead of a pre-defined fixed space.
User memory graph can be a basis for policy because it contains all these valid entities for the current dialog.
In summary, this paper aims to address the following problem:

\noindent \textbf{User Memory Reasoning for Recommendation}:
Assuming an agent involved in a conversational recommendation with a user $e_u$.
The agent 
(1) constructs
\footnote{The construction procedure for user memory graph is omitted here for brevity, and detailed in Section \ref{sec:graph}.} 
a user memory graph
$\mathcal{G}_0 = \{(e, r, e')\vert e, e' \in \mathcal{E}, r \in \mathcal{R} \}$ based on history knowledge $\mathcal{H}$ of $e_u$, candidate items $\mathcal{C}$, and their associated slots and values,
and then, (2) without loss of generality, updates $\mathcal{G}_{x-1}$ with new knowledge from the $x$-th turn $d_x \in D$, in the form of tuples
$\mathcal{G}_x \gets \mathcal{G}_{x-1} \cup \{(e, r, e'), \dots\}$ ;
(3) performs reasoning over $\mathcal{G}_x$ to yield a dialog policy $\pi_x$ that
either (i) performs more rounds of interaction to collect users' knowledge (\eg via question answering), 
or (ii) recommends items $\mathcal{T} \subset \mathcal{C}$ to the user.


To this end, we first collect a dataset for this problem as existing public datasets may hardly meet the needs of this paper for the following reasons.
(1) Lacking users' history and thus dialogs referring to the history (\eg the 2nd and 4th turn in Figure \ref{fig:dialog}).
One reason is that most datasets aim for task-oriented systems, where users' history and reasoning are not core issues to solve.
(2) Lacking fine-grained annotation (for updating the user memory graph). Most public datasets for conversational recommendation are combinations of the datasets for recommender systems and dialogs transcribed separately \cite{li2018towards,zhang2018towards}.
The process is not designed for knowledge-grounded dialogs and leads to the hardness of annotating entity-level knowledge.
(3) Lacking human-level reasoning. The goal of transcribing for existing datasets is not to reason over existing knowledge from both users and items. Some actions are taken at the transcribers' will\cite{li2018towards}.
The collected dataset is called \textbf{M}emory \textbf{G}raph $\leftrightarrow$ \textbf{Conv}ersational \textbf{Rec}ommendation (MGConvRex), containing 7.6K+ dialogs with 73K turns
based on real-world users' behavior. It is annotated with dialog acts, items, slots, values, and sentiment polarities that captures human-level reasoning of dialog policy (see Section \ref{sec:dataset} and Appendix for more details of data collection).

To construct the user memory graph, we define a simple yet flexible ontology, as detailed in Section \ref{sec:graph}.
One challenge in conversational recommendation is to deal with the \textit{open space policy}.
This needs a flexible formation of policy space that differs dialog-by-dialog.
We propose a baseline called user memory graph reasoner (UMGR), which preserves the structure of the user memory graph during reasoning and generates policy based on the graph.  
This also potentially allows for the interpretability of dialog policy.

In summary, the contribution of this paper is as following:
(1) We propose a novel task of user memory reasoning for conversational recommendation;
(2) We collect a dataset and propose an ontology to construct user memory graph;
(3) We propose a baseline for reasoning dialog policy over the user memory graph. 
Experimental results show that such a reasoning model is promising.

\section{Related Work}
\textbf{Conversational Recommendation} is one important type of information seeking dialog system \cite{zhang2018towards}.
Existing studies focus on combining a recommender system with a dialog state tracking system, through the ``System Ask User Respond (SAUR)'' paradigm.
Once enough user preference is collected, such systems often make personalized recommendations to the user.
For instance, \citep{li2018towards} proposes to mitigate cold-start users by learning new users' preferences during conversations and linking the learned preferences to existing similar users in a traditional recommender system.
\citep{sun2018conversational} propose to updates a recommender system in the latent space with the latent space of dialog state tracking and tune the dialog policy via reinforcement learning. 
The updates are short-term and very close to a task-oriented dialog system.
\citep{kang2019recommendation} propose a self-play reinforcement learning (RL) setting to boost the performance of a text-to-text dialog model.
\citep{zhang2018towards} leverages reviews to mimic online conversations to update an existing user's preference and re-rank items.
In \cite{misu-etal-2010-modeling}, the user memory/knowledge is represented as a probabilistic state with a fixed hierarchical structure of Markov probabilistic model to predict dialog actions. However, it lacks the flexibility for encoding richer and fine-grained knowledge and accumulating new knowledge about users for long-term use.
\cite{zhou2020design} demonstrate the usage of user profile and users' interests from ongoing dialog in a social chatbot.
To the best of our knowledge, none of the existing systems (or datasets) aims to build an explicit user memory for reasoning and long-term use.

\noindent \textbf{Task-oriented Dialog Systems} are widely studied with multiple popular benchmark datasets \cite{dstc2, woz, multiwoz, multiwoz2.1,sgd-dst}.
Most of the state-of-the-art approaches \cite{trade,bert-dst-alexa,bert-dst-cmu} focus on improving dialog state tracking with span-based pointer networks for unseen values, which predicts information that is essential for completing a specified task (\eg hotel/air ticket booking, etc.).
Datasets for task-oriented systems typically lack users' history, probably because users' history is not very important to correctly locate a record for the current dialog.
Although certain types of dialog act, slots, and values are shareable for both task-oriented system and conversational recommendation, the core problem of conversational recommendation is to reason and to rank items or questions to ask. 

\noindent \textbf{Graph Reasoning} is essential for generating dialog policy from the proposed user memory graph, where the graph can be viewed as a structured form of state representation. 
There are many studies on leveraging knowledge graphs for recommender systems. For example, \citep{Xian2019ReinforcementKG} introduced a graph-based recommender (not dialog) system that is trained via reinforcement learning.
Graph neural networks are popular in recent years, which aim to learn hidden representations over discrete graph structures\cite{scarselli2008graph,duvenaud2015convolutional,defferrard2016convolutional,kipf2016semi}. 
It is leveraged in this paper to learn structure-preserving (and thus explainable) reasoning.
A number of extensions to the original graph neural network have been proposed \cite{li2015gated,pham2017column},
most notably R-GCNs \cite{schlichtkrull2018modeling}, which can be applied to large-scale and multi-relational graphs (relations are associated with typed embeddings).

A few works have recently been proposed to allow knowledge graph reasoning in dialog systems.
\cite{Moon+19a, Moon+19b} propose a new corpus to learn knowledge graph paths that connect dialog turns.
\cite{tuan-etal-2019-dykgchat} introduces a knowledge-grounded dialog generation task given a knowledge graph that is dynamically updated.
However, these works often focus on response generation and do not address the reasoning of user knowledge in conversational recommendations.

\begin{table*}
    \centering
    \scalebox{0.65}{
        \begin{tabular}{l|l l}
        \hline
        \textbf{Dialog Act $a$} & \textbf{Description} & \textbf{Examples} \\
        \hline
        \textbf{User-side} & & \\
        \hline
        Greeting & Greeting to the agent & I'd like to find a place to eat. \\
        Inform & Actively inform the agent your preference & I'd like to find a \textit{thai} restaurant . \\
        Answer & Answer to a question from the agent & I prefer \textit{thai} food. \\
        Reply & Reply to a recommendation & I'll give it a try.\\
        Open question (OQ) & Actively ask an open question about a recommended item. & What kind of food do they serve ? \\
        Yes/no question (YNQ) & Actively ask an yes/no question about a recommended item. & Do they serve \textit{thai} food ? \\
        Thanks & Thanks the agent & Thanks for your help. \\
        \hline
        \textbf{Agent-side} & & \\
        \hline
        Greeting & Greeting to the user. & How may I help you today ?\\
        Open question (OQ) & Ask an open question about a slot to the user & What kind of food do you prefer ? \\
        Yes/no question (YNQ) & Ask a yes/no question about a value of a slot & I saw you've been to \textit{thai} restaurant, do you still like that ? \\
        Recommendation (REC) & Recommend items to the user. & How about \textit{burger king}, which serves \textit{fast food} ? \\
        Answer (ANS) & Answers user's questions on an item. & They serve \textit{thai} food.\\
        Thanks & Thanks the user & Enjoy your meal. \\
        \hline
        \end{tabular}
    }
    \vspace{-2pt}
    \caption{Dialog acts for agent and user $\mathcal{A}$: the spans of items/slot values are italized.}     
    \vspace{-11pt}
\label{tbl:dialog_act}
\end{table*}

\section{MGConvRex Dataset}
\label{sec:dataset}

This section describes the construction of the \textit{MGConvRex} dataset.
\textit{MGConvRex} aims to contain dialogs that draw relevance of the user's history and fine-grained user preferences to update the user memory graph.
As such, we propose to leverage existing data from recommender systems \footnote{We focus on the restaurant domain at this stage.} that carry users' past behavior to harvest large-scale dialog scenarios.
Then we define fine-grained dialog acts, slots, values and sentiment polarities to turn unstructured utterances into structured knowledge for memory graph updates.

This section is organized as follows.
(1) We detail the curation of dialog scenarios in Section \ref{sec:scenario}.
(2) We then define structured knowledge such as dialog acts, slots, values, and sentiment polarities for MGConvRex, as detailed in Section \ref{sec:semantic}.
(3) Next, we describe the process for transcribing human-to-human simulated dialogs in a Wizard-of-Oz environment \cite{dstc2,woz,multiwoz,multiwoz2.1} (Section \ref{sec:woz}).
(4) Lastly, we define the ontology for annotating the structured knowledge in utterances, and provide the statistics of the dataset in Section \ref{sec:dataset_stat}.
As a result, MGConvRex can be used for a broader scope of research in conversational recommendation, includes but not limited to policy reasoning, natural language understanding (\eg intent detection, slot filling, sentiment analysis), natural language generation, etc.

\subsection{Dialog Scenarios}
\label{sec:scenario}
We use \textit{scenario} to refer to a pre-defined user-agent setting to collect a dialog between two crowd workers, where one plays the user and the other plays the agent. 
Scenarios in conversational recommendation can be generated from user behaviors in the datasets of recommender system.
This mitigates the needs of curating synthetic dialog scenarios as in datasets for task-oriented dialog system\cite{li2016user,li2018microsoft}.

We assume each item is associated with values and each value is associated with at least one slot. 
Let $\mathbb{B}=\{0, 1\}$ be a binary number.
We define a scenario consisting of the following parts: $(e_u, C, H, V, P, \mathcal{T} )$, where $e_u$ is a user, $C \in \mathbb{B}^{\vert \mathcal{C} \vert \times \vert \mathcal{V} \vert}$ is about the candidate items $\mathcal{C}$ and their associated values $\mathcal{V}$, $H \in \mathbb{B}^{\vert \mathcal{H} \vert \times \vert \mathcal{V} \vert}$ is about users past history ($e_u$ visited items $\mathcal{H}$\footnote{To reduce the load of transcribers, a user's past history only contains visited items at this stage.} and their values) that is known to the agent, $V \in \mathbb{B}^{\vert \mathcal{V} \vert \times \vert \mathcal{S} \vert}$ indicates values with their associated slots, $P \in \mathbb{B}^{\vert \mathcal{S} \vert \times \vert \mathcal{V} \vert} $ is the user preference (which value the user prefer for a slot) and $\mathcal{T} \subset \mathcal{C}$ is the ground-truth items. 

We create dialog scenarios as the following way:
(1) for each user, we draw $\vert \mathcal{H} \vert \in [5, 20]$ visited items and $\vert \mathcal{T} \vert = 1$ \footnote{We use 1 ground-truth item to reduce the load of the transcribers and increase the difficulty of reasoning.} items 
as the \textit{ground-truth items} $\mathcal{T}$. Use the values and its associated slots of the ground-truth items as \textit{user preference} $P$.
(2) negatively sample $\vert \mathcal{C} \vert - \vert \mathcal{T} \vert $ items and combine them with the ground-truth items $\mathcal{T}$ as \textit{candidate items} $\mathcal{C}$.

To ensure difficulty of human reasoning, we choose $\vert \mathcal{C} \vert \in [10, 20]$ candidate items and enforce certain similarity over candidate items (such as all locations are from the same state) as the ground-truth items.
For the same user, we also create a duplicated scenario except that $\vert \mathcal{H} \vert = 0$, where the agent player can only use knowledge from the current dialog for recommendation.

\begin{table*}
    \centering
    \scalebox{0.83}{
        \begin{tabular}{l|c|c|c||c|c|c|c}
        \hline
        \textbf{Dataset} & \multicolumn{3}{c||}{\textbf{All Dialogs}} &  \multicolumn{2}{c|}{\textbf{Dialogs w/ History }} & \multicolumn{2}{c}{\textbf{Dialogs w/o History}} \\
        \hline
         & \# of Dial. & \# of Turns & Avg. \# of Turns & \# of Dial. & Avg. \# of Turns & \# of Dial. & Avg. \# of Turns \\
        \hline
Train & 4985 & 48457 & 9.72 & 2418 & 9.62 & 2567 & 9.81 \\
Dev & 263 & 2466 & 9.38 & 121 & 9.16 & 142 & 9.56 \\
Test & 2367 & 23048 & 9.74 & 1160 & 9.62 & 1207 & 9.85 \\
        \hline
        \end{tabular}
    }
    \caption{\textbf{Statistics of the Dataset}: Dialogs w/ or w/o History indicates whether scenarios include visited items $\mathcal{H}$. }     
\label{tbl:dataset}
\end{table*}

\subsection{Dialog Acts, Slots, Values and Sentiment Polarities}
\label{sec:semantic}

We further define the following knowledge for curating structured information for graph updates.

\noindent \textbf{Dialog Acts} ($\mathcal{A}$): Table \ref{tbl:dialog_act} demonstrates the dialog acts for both the user and the agent.
Note that besides the System Ask – User Respond (SAUR) paradigm \cite{sun2018conversational,li2018towards,zhang2018towards}, we also propose a User Ask - System Respond (UASR) paradigm that allows users to actively participate in a recommendation. 
Acts such as \textit{Open question}, \textit{Yes/no question} and \textit{Inform} are designed for this purpose.

\noindent \textbf{Slots and Values}($\mathcal{S}$, $\mathcal{V}$): 
We select $\vert \mathcal{S} \vert = 10$ popular slots with a total of 470+ values for the restaurant domain.
To help transcribers use some values naturally in utterances, we change some values (such as price ranges \$) into English words (``cheap'' etc.).

\noindent \textbf{Sentiment Polarity}: We define a user's preference expressed in a conversation as 
pairs of opinion targets (an item or a value) and their associated sentiment polarities\cite{hu2004mining}.
We adopt 3 types of polarities \textit{pos\_on}, \textit{neg\_on} and \textit{neu\_on} to represent positive, negative and neutral polarity, respectively
\footnote{We do not deal with emotions (\eg \textit{sad}), although existing works may use sentiment to indicate emotions.}.


\subsection{Wizard-of-Oz Collection and Annotation}
\label{sec:woz}

We build a wizard-of-oz system to randomly pair two crowd workers to engage in a chat session, where each scenario is split into two parts:
$(P, \mathcal{T})$ for the user and $(e_u, C, H, V)$ for the agent.
The goal of a conversation is like a game between the user and the agent, where the agent needs to reason the user's current preference and find the ground-truth item and the user can tell information from preference $P$ or confirm a recommended item $e_i \in \mathcal{T}$ but cannot tell the ground-truth directly. 
The guidelines, screenshots of the Wizard-of-Oz UI can be found in the Appendix.

\subsection{Summary of MGConvRex}
\label{sec:dataset_stat}
We annotate dialog acts, items, slots, values, and users' utterance-level and entity-level sentiment.
The dialogs are split into training, development, and testing sets with non-overlapping users for 
zero-shot reasoning on unseen users.
The statistics of MGConvRex are in Table \ref{tbl:dataset}.
For scenarios with users' history, we notice that the average number of turns are slightly shorter than those without users' history.
We further plot agent's dialog acts to study the behavior of the agent players, as in Figure \ref{tbl:dialog_act}, where agent players seem to use more yes/no questions to confirm users' preference exhibit in history.
We discuss more details in Appendix.

\begin{figure}[t]
\centering    
\includegraphics[width=3.2in]{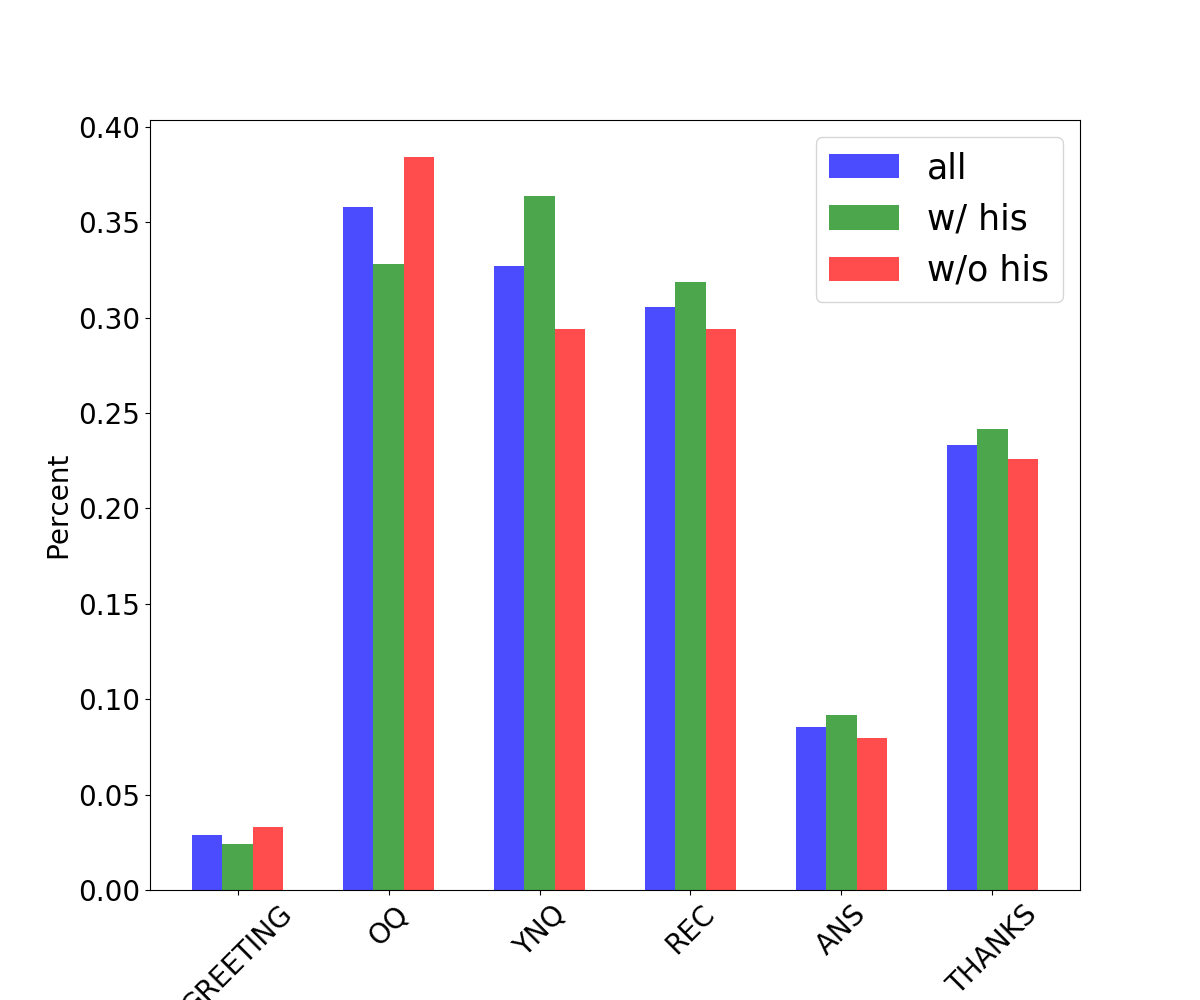}
    \caption{Distribution of dialog acts from agent side: w/ his indicates scenarios have users' history.}
\label{fig:agent_act}
\end{figure}

\section{User Memory Graph}
\label{sec:graph}

In this section, we describe the formulation of a user memory graph based on a scenario and annotated user preference.
There are many design choices for constructing a user memory graph.
Our goal is to model user knowledge and scenarios with extensibility and maintenance.

\begin{table}
    \centering
    \scalebox{0.8}{
        \begin{tabular}{c|l}
        \hline
        \textbf{Entity Types} $\mathcal{E}$ & \textbf{Explanation} \\
        \hline
        $\mathcal{U}$ & user entities \\
        $\mathcal{M}$ & memory entities\\
        $\mathcal{I}$ & item entities: $\mathcal{C} \cup \mathcal{H}$\\
        $\mathcal{S}$ & slot entities \\ 
        $\mathcal{V}$ & value entities \\
        \hline
        \textbf{Relation Types} $\mathcal{R}$ &   \\
        \hline
        $(\mathcal{U}, \text{has\_memory}, \mathcal{M})$ & a user $u$ has a memory entity $m$\\
        $(\mathcal{M}, \text{visited}, \mathcal{I})$ & a memory $m$ is about an item $i$\\
        $(\mathcal{I}, \text{has\_aspect}, \mathcal{V})$ & an item $i$ has a value $v$\\
        $(\mathcal{V}, \text{is\_a}, \mathcal{S})$ & a value $v$ belongs to a slot $s$\\
        $(\mathcal{M}, \textbf{pos\_on}, \mathcal{V}/\mathcal{I})$ & $m$ is positive on a value or item\\
        $(\mathcal{M}, \textbf{neg\_on}, \mathcal{V}/\mathcal{I})$ & $m$ is negative on a value or item\\
        $(\mathcal{M}, \textbf{neu\_on}, \mathcal{V}/\mathcal{I})$ & $m$ is neutral on a value or item\\
        \hline        
        \end{tabular}
    }
    \vspace{-2pt}
    \caption{Ontology of user memory graph: bolded relations are used for graph updates or accumulation.} 
    \vspace{-10pt}
\label{tbl:meta_entity_edges}
\end{table}

\begin{figure*}[t]
\centering    
    \begin{subfigure}[b]{0.3\textwidth}
    \centering
    \includegraphics[width=\textwidth]{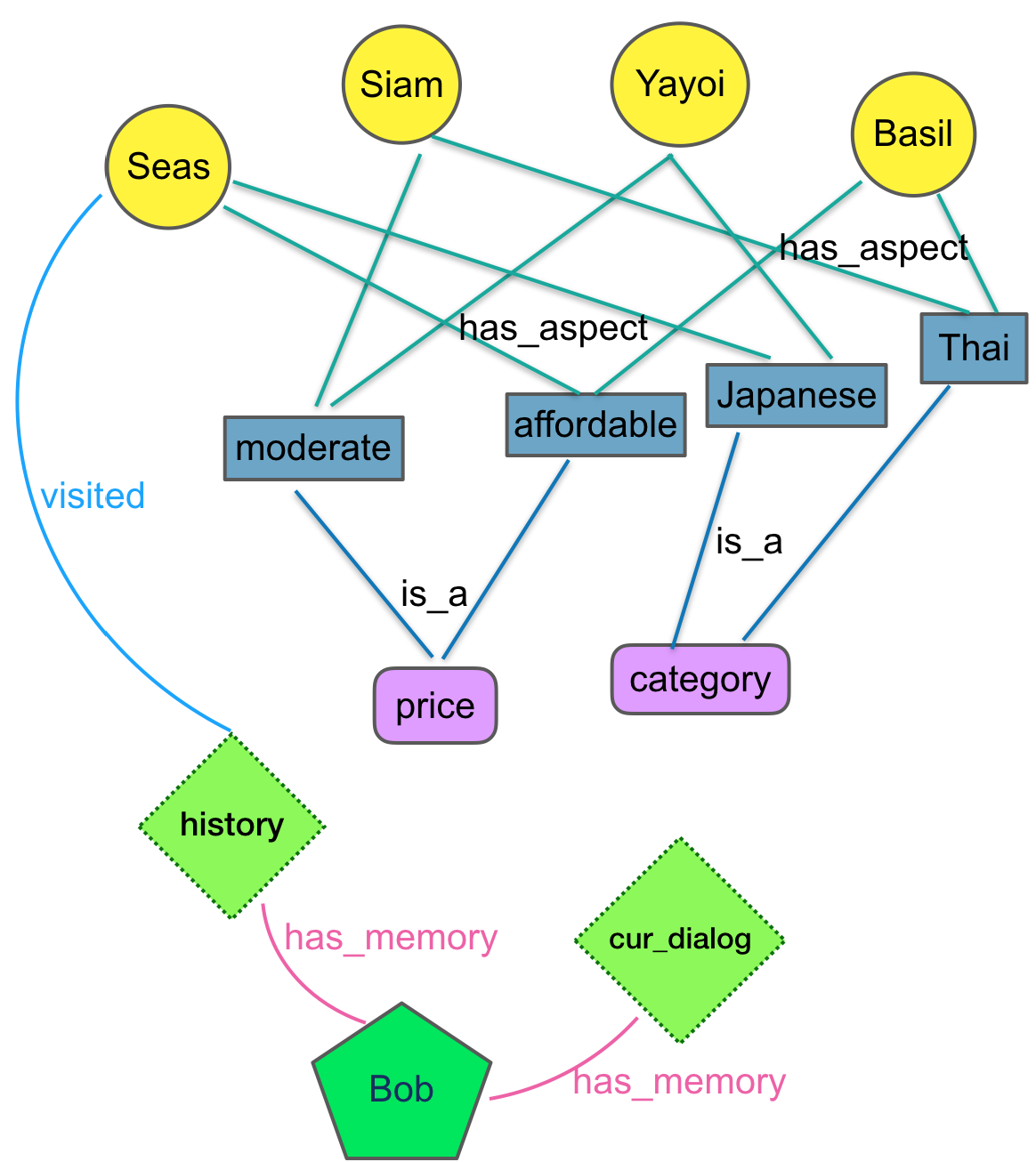}
    \end{subfigure}
    \hfill
    \begin{subfigure}[b]{0.3\textwidth}
    \centering
    \includegraphics[width=\textwidth]{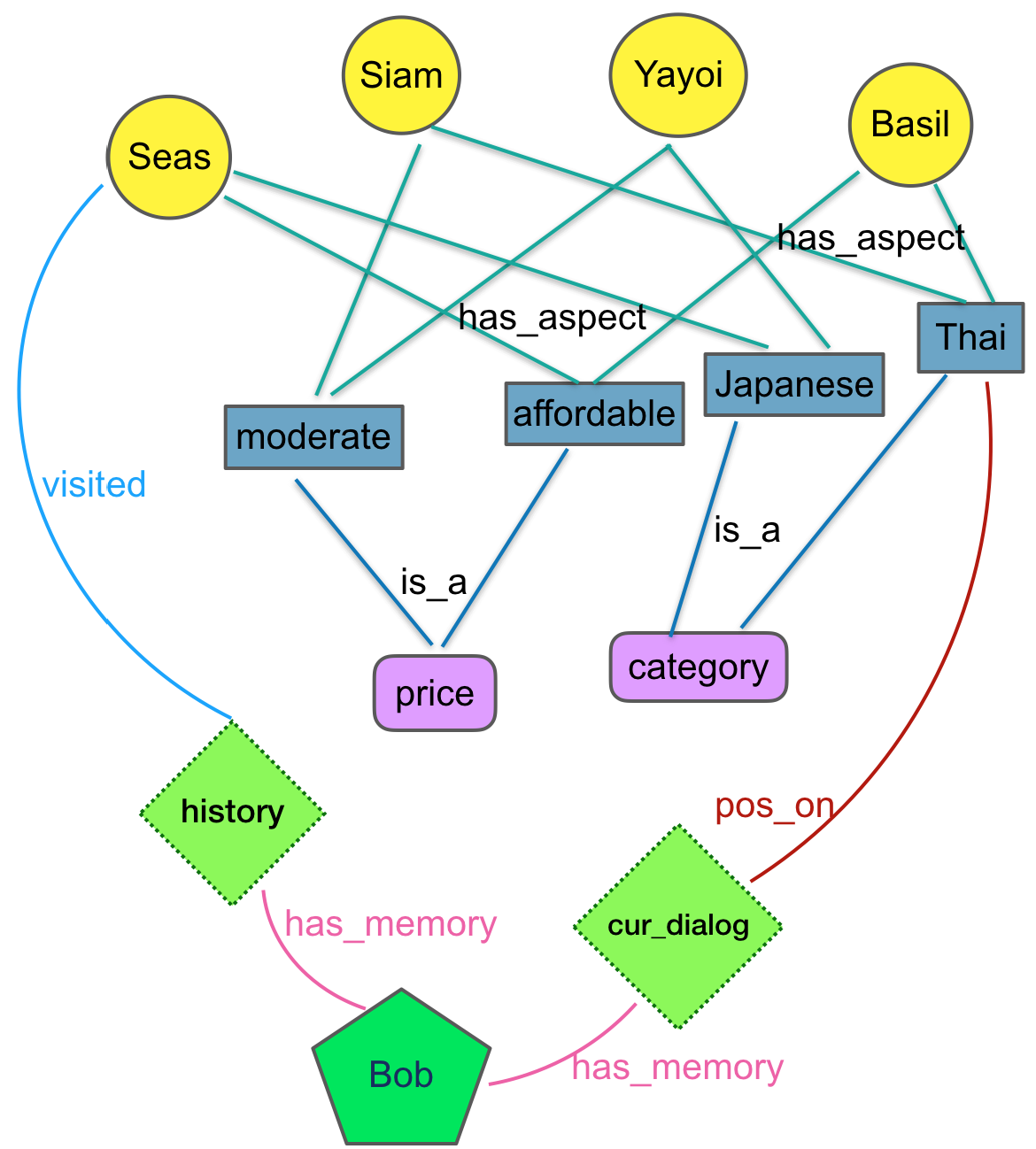}
    \end{subfigure}
    \hfill
    \begin{subfigure}[b]{0.3\textwidth}
    \centering
    \includegraphics[width=\textwidth]{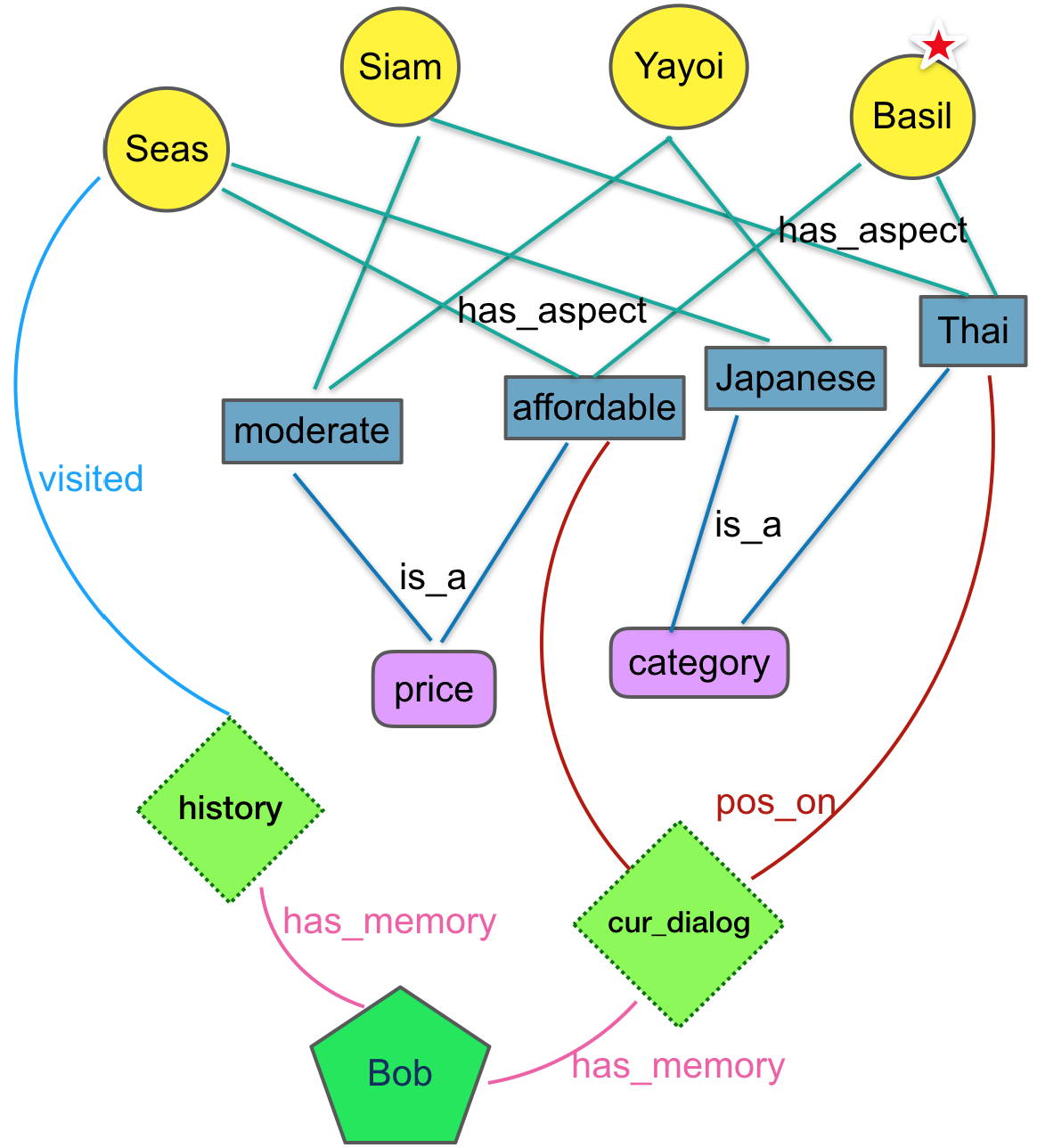}
    \end{subfigure}
\caption{User memory graph construction and updates based on the dialog in Table \ref{tbl:chat}.}
\label{fig:cum}
\vspace{-3mm}
\end{figure*}

\subsection{Construction}
As a reminder, a user memory graph is denoted as $\mathcal{G} = \{(e, r, e')\vert e, e' \in \mathcal{E}, r \in \mathcal{R} \}$, which is essentially a heterogeneous graph with typed entities and relations.
We first define the ontology (or meta entities and relations) in Table \ref{tbl:meta_entity_edges}.
The user memory contains available items $\mathcal{I}$ for a dialog scenario.
An item $i$ can be associated with multiple values $v$s with $r_\text{has\_aspect}$ relation.
Each value is associated with their slot $s$ via $r_\text{is\_a}$ relation.
In this way, values / slots entities are rather expandable and new values or slots (or even slots of slots) can be easily added in.
Further, each user has their own entity $e_u$ and several associated memory entities $m$s.
We define memory entity to model an event or experience of the user, such as visiting a restaurant (via entity $m_\text{history}$), or having a conversation as in current dialog (via $m_\text{cur\_dialog}$).
The advantage of allowing multiple memory entities is that a user may have different opinions for the same target (items or values) from their very different experiences (\eg like \textit{Thai} food for lunch but not dinner). 
To express a user's history on visited items, we use a $r_\text{visited}$ relation to connect a memory entity with a visited item.
As an example, we demonstrate the construction of a user $u_\text{Bob}$ in the first graph in Figure \ref{fig:cum}. 
We will keep use this example to demonstrate the updates of user memory graph for the dialog in Table \ref{tbl:chat}.

\subsection{Update}
The updates of user memory graph is assumed\footnote{We leave language understanding parts to future work and the baselines of this paper use ground-truths from annotations.} to leverage the outputs of natural language understanding (NLU) or state tracking.
For simplicity, we use 3 sentiment relations $r_\text{pos\_on}$, $r_\text{neg\_on}$ and $r_\text{neu\_on}$ to update a user memory graph, which associate values/items (opinion target) with the memory entity of the current dialog $m_\text{cur\_dialog}$.
We believe humans have a more complex memory system in their brains. We expect more complex (such as error correction) memory update systems in future work.

From the first turn of the user in Table \ref{tbl:chat}, we know that $u_\text{Bob}$ likes \textit{Thai} food and the user memory graph is updated with a new triple $(m_\text{cur\_dialog}, r_\text{pos\_on}, v_\text{Thai})$.
Following the second turn of the user, we know that $u_\text{Bob}$ is still interested in $v_\text{affordable}$ restaurants, indicated by a new triple $(m_\text{cur\_dialog}, r_\text{pos\_on}, v_\text{affordable})$.
Then the agent can infer a recommendation $i_{\text{Basil}}$, which can be explained by paths: 
(1) $u_\text{Bob} \rightarrow r_\text{has\_memory} \rightarrow m_\text{cur\_dialog} \rightarrow r_\text{pos\_on} \rightarrow v_\text{Thai} \rightarrow r_\text{has\_aspect} \rightarrow i_\text{Basil}$, 
(2) $u_\text{Bob} \rightarrow r_\text{has\_memory} \rightarrow m_\text{cur\_dialog} \rightarrow r_\text{pos\_on} \rightarrow v_\text{affordable} \rightarrow r_\text{has\_aspect} \rightarrow i_\text{Basil}$, 
and (3) $ u_\text{Bob} \rightarrow r_\text{has\_memory} \rightarrow m_\text{history} \rightarrow r_\text{visited} \rightarrow v_\text{Seas} \rightarrow r_\text{has\_aspect} \rightarrow v_\text{affordable} \rightarrow r_\text{has\_aspect} \rightarrow i_\text{Basil}$, where the last path draws the relevance from a visited item to the current recommendation. 
As we can see, sentiment relations serve as the bridge to connect a user to items and enables potential reasoning for recommendation.

\begin{table}
    \centering
    \scalebox{0.7}{
        \begin{tabular}{c|l}
        \hline
        \textbf{Role} & \textbf{Utterance} \\
        \hline
        Agent & what kinds of food do you like ?\\
        User & I like \textbf{Thai} food.\\
        Agent & are you \textit{still} interested in \textbf{affordable} restaurant ?\\
        User & \textit{yes}.\\
        Agent & how about \textbf{Basil}, which is affordable and serves Thai food.\\
        \hline        
        \end{tabular}
    }
    \vspace{-2pt}
    \caption{An example dialog corresponds to the graph updates in Figure \ref{fig:cum}.} 
    \vspace{-10pt}
\label{tbl:chat}
\end{table}

\section{User Memory Graph Reasoner}
\label{sec:umgr}
\begin{figure}[t]
\centering    
\includegraphics[width=3.in]{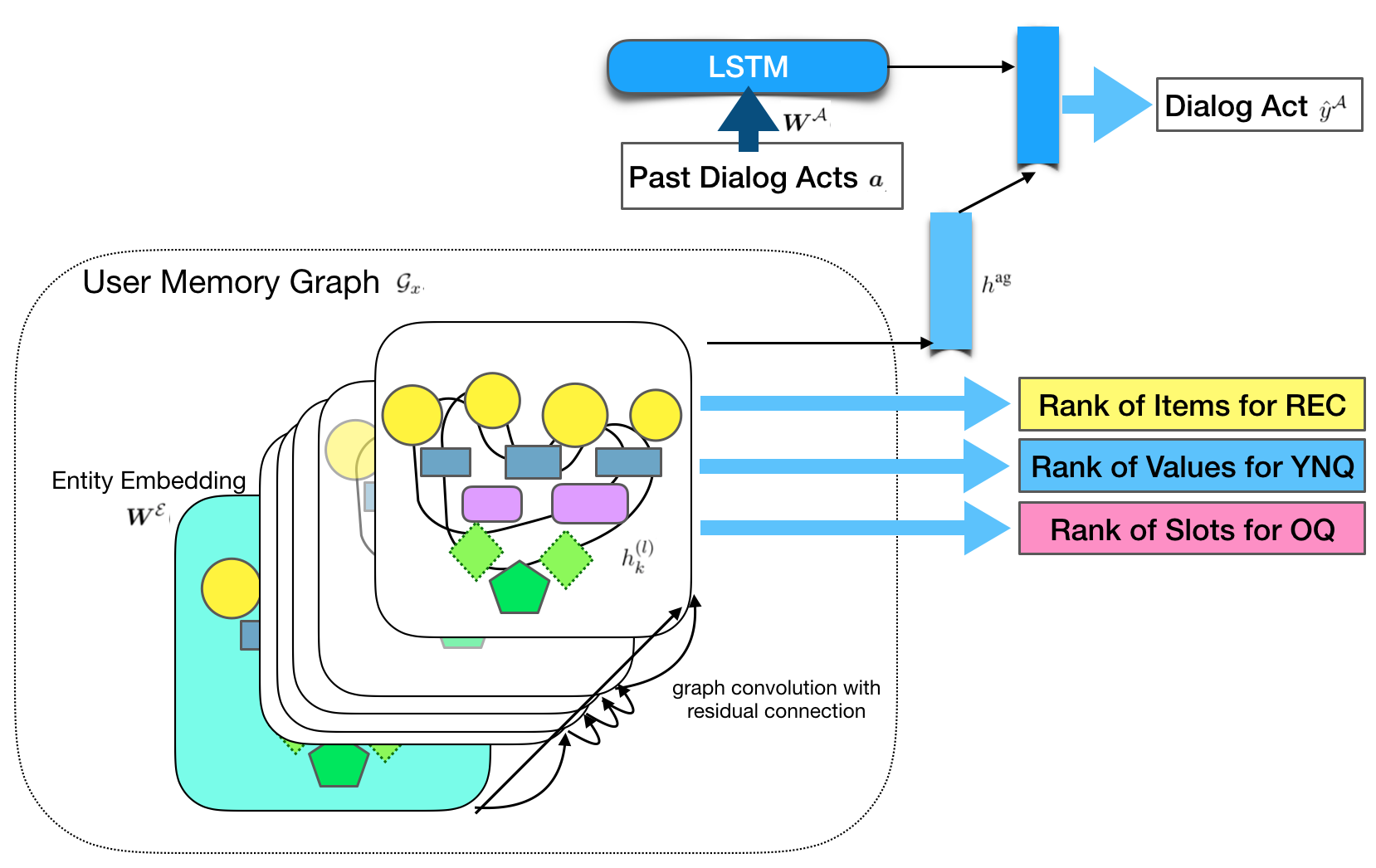}
    \caption{Overview of the User Memory Graph Reasoner (UMGR) architecture.}
\label{fig:model_overview}
\end{figure}

In this section, we propose a model called User Memory Graph Reasoner (UMGR), which uses user memory graph to reason dialog policy (Figure \ref{fig:model_overview}). 
As discussed in the introduction, we aim to resolve the issue of open space policy in conversational recommendation. 
We define the inputs/outputs as following, which maps certain entities from user memory graph to policy space.

\noindent \textbf{Input}: (1) past dialog acts up to the current turn from the user $\boldsymbol{a}$; (2) updated user memory graph $\mathcal{G}_x$.

\noindent \textbf{Output}: dialog policy $\pi=(\hat{y}^\mathcal{A}, \hat{y}^\mathcal{C}, \hat{y}^\mathcal{S}, \hat{y}^\mathcal{V})$ for the current turn, where $\mathcal{A}$, $\mathcal{C}$, $\mathcal{S}$, $\mathcal{V}$ indicate the space of dialog acts, candidate items, slots and values, respectively.

Note that $\hat{y}^\mathcal{C}, \hat{y}^\mathcal{S}$ and $\hat{y}^\mathcal{V}$ can be interpreted as the arguments of dialog acts and are essentially rankings over their corresponding entity sets.
For example, when $\hat{y}^\mathcal{A}=$ \textit{Recommendation}, the top-1 entity $\argmax_{e_i \in \mathcal{C}}(\hat{y}^\mathcal{C})$ will be provided to the user.
Similarly, $\hat{y}^\mathcal{A}=$ \textit{Open Question} is related to the top-1 slot $\argmax_{e_s \in \mathcal{S}}(\hat{y}^\mathcal{S})$ and $\hat{y}^\mathcal{A}=$ \textit{Yes/no Question} is related to the top-1 value $\argmax_{e_v \in \mathcal{V}}(\hat{y}^\mathcal{V})$.
As such, the policy space of UMGR can be determined by the user memory graph where only valid entities can be generated. A structure-preserving model is preferred for reasoning where all entities in policy are generated as a holistic reasoning process.

We let UMGR first encodes past dialog acts $\boldsymbol{a}$ and entities $e_j \in \mathcal{E}$ into hidden dimensions.
\begin{equation}
\begin{split}
h_a = \text{LSTM}(\boldsymbol{W}^\mathcal{A}(\boldsymbol{a})), \\
h_j^{(0)} = \boldsymbol{W}^\mathcal{E}(e_j),
\end{split}
\end{equation}
where $\boldsymbol{W}^\mathcal{A}$ and $\boldsymbol{W}^\mathcal{E}$ are embedding layers and the past dialog acts are further encoded by an LSTM encoder.
Then we incorporate a Relational Graph Convolutional Networks (R-GCN) \cite{schlichtkrull2018modeling} into UMGR for reasoning.
R-GCN is a GCN \cite{kipf2016semi} with typed relations, where each relation is associated with their own weights to enable reasoning over a heterogeneous graph.
Each entity is encoded by multiple layers of R-GCN as following:
\begin{equation}
\begin{split}
h_j^{'(l+1)}=\text{GELU} \Big(\sum_{r \in \mathcal{R}} \sum_{k \in \mathcal{N}_j^r} \frac{1}{\vert \mathcal{N}_j^r \vert} \boldsymbol{W}_r^{(l)} h_k^{(l)}\Big),
\end{split}
\end{equation}
where $h_j^{(l)}$  is the hidden state of entity $e_j$ in the $l$-th layer of R-GCN, $\mathcal{N}_j^r$ is entity $e_j$'s neighbors in relation type $r$ and $\boldsymbol{W}_r^{(l)}$ is the weight associated with $r$ in the $l$-th layer to transform one neighbor $h_k^{(l)}$.
The R-GCN layer updates the hidden states of each entity with the incoming messages in the form of their neighbors' hidden states type-by-type.
Then R-GCN sums over all types before passing through the GELU activation \cite{hendrycks2016gaussian}.
The hidden state of entity $e_j$ in the $(l+1)$-th layer is computed via a residual connection \cite{he2016deep} (to keep the original entity information instead of just neighbors' information) and layer normalization.
\begin{equation}
\begin{split}
h_j^{(l+1)}=\text{LayerNorm}\Big(h_j^{(l)} + h_j^{'(l+1)}\Big).
\end{split}
\end{equation}

The hidden states from the last layer of R-GCN is passed into an aggregation layer.
\begin{equation}
\begin{split}
h^{\text{ag}} = \frac{1}{\vert \mathcal{C} \cup \mathcal{S} \cup \mathcal{V} \vert} \sum_{e_j \in \mathcal{C} \cup \mathcal{S} \cup \mathcal{V}} (\boldsymbol{W}^\text{ag} h_j^{(l+1)} + b^{\text{ag}}),
\end{split}
\end{equation}
where $\boldsymbol{W}^{\text{ag}}$ and $b^{\text{ag}}$ are weight for aggregation layer.
The purpose of having an aggregation layer is to leverage the information in the user memory graph for predicting the dialog acts.
The loss for dialog acts is defined as
\begin{equation}
\begin{split}
\hat{y}^{\mathcal{A}} = \text{Softmax}\Big(\text{MLP}^{\mathcal{A}}(\boldsymbol{W}^\mathcal{A} (h_a \oplus h^\text{ag}) +b^\mathcal{A}) \Big), \\
\mathcal{L}^{\mathcal{A}} = \text{CrossEntropyLoss}(\hat{y}^\mathcal{A}, y^\mathcal{A}),
\end{split}
\end{equation}
where $\oplus$ is the concatenation operation, $\boldsymbol{W}^\mathcal{A}$ merges the hidden states of dialog acts and graph, $\text{MLP}^{\mathcal{A}}(\cdot)$ is a multi-layer perception for dialog acts and $y^\mathcal{A}$ is the label of dialog act.
Further, all item, slot and value entities are trained by log loss for ranking. For example, the loss for candidate items $\mathcal{C}$ is computed as 
\begin{equation}
\begin{split}
\hat{y}_i = \text{Sigmoid}\Big(\text{MLP}^\mathcal{I}(h_i) \Big), \\
\mathcal{L}^\mathcal{C} = \text{LogLoss}(\hat{y}^\mathcal{C}, y^\mathcal{C}),
\end{split}
\end{equation}
where $\text{MLP}^\mathcal{I}(\cdot)$ is the multi-layer perceptron for item.
Similarly, we obtain loses $\mathcal{L}_\mathcal{S}$, $\mathcal{L}_\mathcal{V}$ for slot entities $\mathcal{S}$ and value entities $\mathcal{V}$, respectively.
The total loss is the sum over all losses for dialog acts, items, slots and values: 
\begin{equation}
\begin{split}
\mathcal{L} = \alpha\mathcal{L}^\mathcal{A} + \beta\mathcal{L}^\mathcal{C} + \gamma\mathcal{L}^\mathcal{S} + \delta\mathcal{L}^\mathcal{V},
\end{split}
\end{equation}
where $\alpha, \beta$, $\gamma$ and $\delta$ are hyper-parameters to balance losses of different scales.
Note that during training and prediction, all invalid entities (\eg not appear in a user memory graph) are masked out.
As we can see, unlike traditional recommender systems, UMGR has no assumption on users/items in training set and provides the capability of zero-shot reasoning. The policy space is open-ended because entities in policy is determined by the rankings of entities in user memory graph instead of a pre-defined set for the model.

\section{Experiments}
\label{sec:exp}
This section conducts experiments on baselines for reasoning dialog policy.

\subsection{Evaluation Metrics}
We propose the following metrics to evaluate UMGR both offline (against the collected testing dialogs) and online (against a user simulator running on testing scenarios in MGConvRex.

\subsubsection{Offline metrics}
We propose the following offline metrics to evaluate UMGR. Note that all offline metrics assume UMGR uses annotations (ground-truth) of past turns (\eg on constructing a user memory graph).
\noindent \textbf{Act Accuracy \& F1} are reported for all predicted dialog acts against annotated turn acts in testing.

\noindent \textbf{Entity Matching Rate (EMR, k@1, 3, 5)} measures turn-level top-$k$ entities against the testing set. 
These metrics evaluate only on correctly predicted dialog acts since the types of predicted entities (items, slots, or values) depend on the predicted dialog acts $\hat{y}^\mathcal{A}$.

\noindent \textbf{Item Matching Rate (IMR)} measures dialog-level predicted items against the ground-truth items.

\subsubsection{Online metrics}
In addition to offline evaluation, we use a user simulator (see Appendix) to dynamically evaluate the performance of recommendation. This mitigates the assumption in offline metrics that all past turns are correct, which limits the interactive evaluation of conversations.

\noindent \textbf{Success Rate} tracks whether the interaction with user simulators yields the ground-truth item $e_t$. We use the scenarios for testing sets used for the offline evaluation. The maximum number of turns is simulated as 11. We ran simulations \textbf{3} times and average the results.

\begin{table*}[!t]
    \centering
    \scalebox{0.83}{
        \begin{tabular}{l|l|l|l|l|l|l||c}
        \hline
\multirow{3}{*}{\textbf{Methods}} & \multicolumn{6}{c||}{\textbf{Offline Evaluation}} & \multicolumn{1}{c}{\textbf{Online Evaluation}} \\
\hline
& \textbf{Act Acc.}  & \textbf{Act F1} & \multicolumn{3}{c|}{\textbf{EMR}} & 
\textbf{IMR} & \textbf{Success Rate} \\
\hline
& & & @1 & @3 & @5 &  & \\
\hline
\hline

RandomAgent & 18.17 & 18.24 & 1.5 & 1.5 & 1.5 & 6.55 & 6.0 \\
RecAgent      & 25.89 & 6.86 & 2.7 & 2.7 & 2.7 & 39.16 & 39.21 \\
Pretrained Emb. & 64.3 & 54.79 & 13.75 & 29.02 & 36.7 & 9.97 & 9.73 \\
MemoryNetwork & 59.46 & 53.78 & 13.85 & 29.46 & 35.82 & 4.73 & 6.31 \\
\hline
\hline
\textbf{UMGR (Proposed)} & \textbf{65.7} & \textbf{56.54} & \textbf{33.92} & \textbf{48.47} & \textbf{52.54} & 67.93 & \textbf{71.03} \\
- Prev. User Act Only & 63.47 & 54.64 & 33.66 & 46.69 & 50.59 & \textbf{69.71} & 69.76 \\
- No Dialog Acts & 42.37 & 32.72 & 31.52 & 43.66 & 46.89 & 67.6 & 66.1 \\
- Static $\mathcal{G}$ & 64.31 & 55.25 & 18.03 & 36.9 & 45.31 & 27.5 & 37.26 \\

\hline
        \end{tabular}
    }
    \vspace{-2pt}
    \caption{Results of both offline and online evaluation: EMR stands for entity matching rate, which compares all types of predicted entities against annotated ones when the dialog act is predicted correctly; IMR stands for item matching rate, which evaluates predicted items against the ground-truth item across all turns in a dialog.}     
    \vspace{-2pt}
\label{tbl:result}
\end{table*}

\begin{figure*}[t]
\centering    
\includegraphics[width=6.3in]{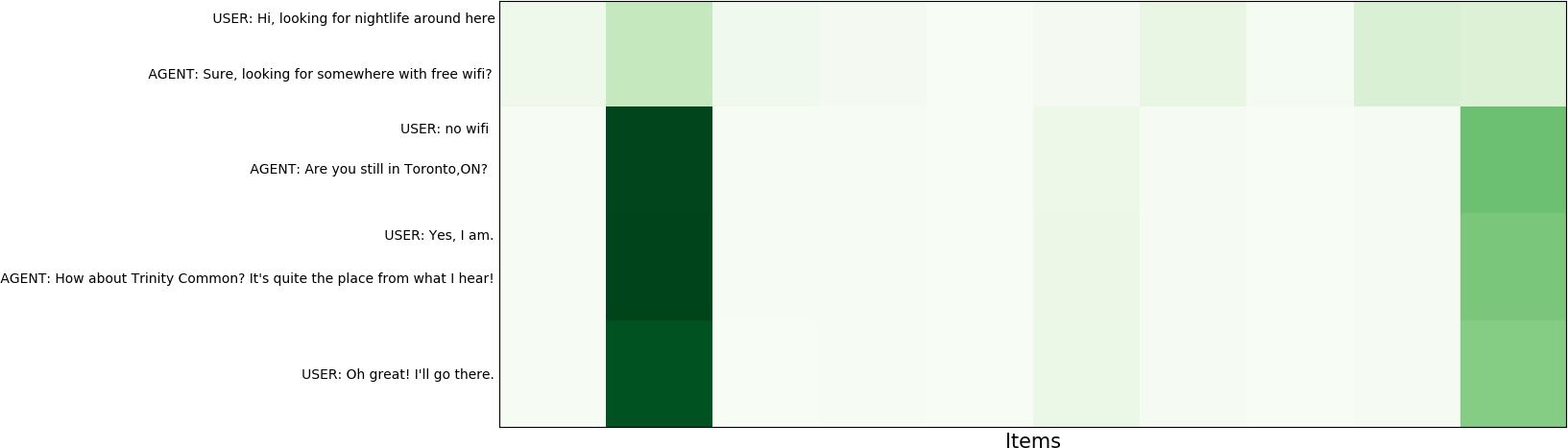}
    \caption{Visualization of item-level conversational reasoning, given an example dialog. Darker color indicates more salient items for recommendation at each given turn (row), predicted by our UMGR model.}
\label{fig:ranking}
\vspace{-3mm}
\end{figure*}

\subsection{Compared Methods}
\noindent \textbf{RandomAgent}: we implement a baseline agent that randomly picks a dialog act and randomly pick a candidate item/slot/value as the dialog policy.

\noindent \textbf{RecAgent}: this agent always chooses \textit{Recommendation} as the optimal dialog act to enact and select a random item that has not been tried in candidate items (memorize all trials). This is a strong (yet annoying) rule-based baseline and does not collect or use any user preference.

\noindent \textbf{Memory Network}\cite{sukhbaatar2015end,bordes2016learning}: we adapt memory network and encodes the user memory graph as triples. The memory can be updated as new triples added. Note that memory networks cannot deal with open space policy because of attention-based aggregation of triple memories. As such, we enumerate all possible combinations of dialog acts and entities in user memory as the space of policy. Specifically, all items in a scenario are indexed as $i_1, i_2, \dots$ to differentiate candidate items for policy generation. The inputs of the memory network are the encoded dialog acts (the same as UMGR). We adopt 5 hops for memory networks.

\noindent \textbf{Pretrained Embeddings}: 
we pre-train the graph embeddings and utilize these as graph encoder for predicting dialog policy (without  R-GCN layers in UMGR).
The graph embeddings are trained from all scenarios in the training set using the TransE-based graph prediction approaches \cite{Nickel+15}.
While this approach is widely used in the related literature and carries cross-scenario knowledge, we show that using pre-trained graph embedding alone is sub-optimal for a particular user's scenario and the dialog policy needs to perform dynamic reasoning over the user memory graph.

\noindent \textbf{UMGR} (\textbf{Proposed}): this is the proposed model in Section \ref{sec:umgr}. 
To enable zero-shot reasoning during inference, all items share the same embeddings and UMGR purely learns leverage the graph structure for reasoning policy.
We adopt 5 layers of R-GCN and all sizes of hidden states are 384. The maximum number of past acts is set as 10.
Factors of losses $\alpha$, $\beta$, $\gamma$ and $\delta$ are set as 1, 10, 10, 100 based on the scales of losses. 
We choose the batch size to be 160.
We further investigate the following ablation studies on UMGR:

\noindent \textbf{- Prev. User Act Only}: this ablation study only uses the most recent dialog act from the user. We use this to show how many past dialog acts are needed for policy generation.

\noindent \textbf{- No Dialog Acts}: this study removes the dialog acts encoder, investigating the importance of the dialog acts for recommendation.

\noindent \textbf{- Static $\mathcal{G}$}: this study uses the initial user memory graph without any updates during the conversation. We use this study to demonstrate that dynamic updates of the user memory graph are crucial for reasoning better dialog policy.

\subsection{Results and Discussion}
The results are summarized in Table \ref{tbl:result}.
Overall, it can be seen that the proposed UMGR architecture outperforms other baselines in both offline and online evaluation.
\textbf{Ablations}: Specifically, we notice that dynamically updating the user memory graph with users' new preference is crucial for a recommendation, as indicated by \textit{UMGR - static $\mathcal{G}$} that forbids updating user memory graph. 
It can also be seen that removing the previous dialog context does degrade the performance as expected (\textit{UMGR - Prev. User Act Only}), although the UMGR architecture still maintains a competitive performance. 
Similarly, while \textit{UMGR -No Dialog Acts} does not take past dialog acts as input, its results on non-act prediction metrics are relatively competitive.
Both of these ablation studies indicate the user memory graph contains enough information for the model to perform dialog reasoning.

\noindent \textbf{UMGR vs. Memory Network}. We notice that memory networks may not be suitable for complex reasoning over a user memory graph. This may be caused by the following reasons: (1) triples in memory are disconnected, which limits the possibility of joint reasoning of multiple triples; (2) memory network is not structure-preserving, which leads to hardness of aligning entities in triples with the output policy, such as ranking items; (3) existing research using memory network \cite{bordes2016learning,eric2017key,madotto2018mem2seq} assumed a static memory, which carries a great amount of knowledge from training to testing. Memory network may not be very suitable for our zero-shot reasoning where no user or item knowledge can be carried to testing directly.

\noindent \textbf{UMGR vs. Rule-based Agent}. We notice that \textit{RecAgent} is a good rule-based baseline regarding the performance of recommendation. One advantage of RecAgent is that it can easily remember the recommended items tried in previous turns. However, frequent acts of recommendation can be annoying to the user.

\noindent \textbf{UMGR vs. Pre-trained Graph Embeddings}. We confirm that static pre-trained graph embeddings provide general representations of memory graphs but have a limited capability of reasoning for a particular user's scenario. This study indicates UMGR has the capability for a personalized recommendation.

\noindent \textbf{Discussion} 
We first examine the generated dialog acts. 
UMGR typically asks a few questions and then makes a few recommendations. 
We observe that UMGR may make more recommendations than expected from agent workers in MGConvRex.
This may be caused by the frequent patterns of dialog acts in conversational recommendation: different types of non-recommendation acts are frequently followed by a recommendation act.
As a result, a neural network prefers frequent patterns to diverse details of reasoning.
We believe more diverse and detailed reasoning is an important direction to improve in the future.
Meanwhile, we argue that human performance on reasoning is very limited given the vast amount of candidate items in the real-world recommendation.
Learning the behavior from humans is just a beginning. We expect research on automatic reasoning over large-scale user knowledge in future work.

\noindent \textbf{Visualization of Item-level Reasoning}.
Figure \ref{fig:ranking} shows an example dialog in which the prominence scores of candidate items for recommendations at each turn, predicted by our model (darker color indicates more salient items for recommendation).
At the beginning of the dialog, the prominence scores (and thus the ranking among the candidate items) are soft-initialized to reflect the user's offline preferences, as indicated in the user memory graph. 
We can see that UMGR can almost predict the ground-truth item. 
As the dialog progresses and the system collects (or confirms) new user knowledge or a request (\eg updated slots, opinions on recommended items ``Toronto,ON", \etc), the proposed UMGR model dynamically updates the ranking of the relevant items, reflecting the online preferences.
Overall, UMGR effectively incorporates both online and offline preferences through a structured user memory graph, allowing for natural interactions and accurate recommendations.

\section{Conclusion}
This paper proposes a novel problem of user memory graph reasoning for conversational recommendation.
We expect to release a conversational recommendation dataset with a grounded user memory graph from the behaviors of real-world users.
The proposed user memory graph has the benefits of accumulating knowledge for a user to reason dialog policy.
We propose a baseline model called UMGR that performs reasoning over such a user memory graph in open space policy.
UMGR is structure-preserving for policy generation and provides zero-shot reasoning capability for user memory graphs that have never been seen before. 
Experimental results demonstrate the effectiveness of UMGR over a wide spectrum of metrics.

\section*{Acknowledgments}
We thank Rajen Subba, Alborz Geramifard, and Hao Zhou for insightful discussions. 
Thanks to Gerald Demeunynck for the discussion and improvement on the process of data collection.

\bibliography{anthology,acl2020}
\bibliographystyle{acl_natbib}

\clearpage
\appendix
\section{Appendix}
This appendix contains two guidelines for building MGConvRex dataset: transcription guideline and annotation guideline, followed by the statistics of the dataset and a sample implementation of user simulator.

\subsection{Transcription Guideline}
\label{sec:transcrib}
\subsubsection{Motivation}
Getting irrelevant restaurant recommendations is a frustrating experience.
The ideal recommendation system should be able to provide better recommendations by understanding your current needs, your restaurant preferences, and your restaurant history.

\subsubsection{Overview}
In this project, you will generate a dialog between an imaginary person (\textit{user}) and an imaginary recommendation system (\textit{assistant}\footnote{We term agent as ``assistant'' in guidelines.}).  You will play one of the two roles, that will randomly be assigned to you. You will automatically get paired with someone else who will play the other role.

\noindent \textbf{User}: A user is expected to interact with an assistant to get a restaurant recommendation.  The user will already know his/her general restaurant preferences and also the exact name of the restaurant he/she wants to go to. Further, information about restaurants that the user has visited in the past will be available and shown to the user.

\noindent \textbf{Assistant}: An assistant is expected to interact with the user and work towards recommending a restaurant the user wants to go to in the future. The assistant will have access to information about restaurants that the user has previously visited and a list of candidate restaurants.

\subsubsection{Task}
You will be randomly assigned a single role: either \textit{user} or \textit{assistant}. You will see your assignment in the top left corner of the screen, ``You are: the user'' or ``You are: the assistant''.

\noindent \textbf{User}: You will interact with the assistant, to get the correct restaurant recommendation from the assistant. You will be provided with the following information: 
\begin{itemize}
\itemsep0em
    \item Restaurant preference over 10 characteristics (or slots).
    \item The restaurant you will go to: ``Ground-Truth restaurant''.
    \item You will optionally have information about restaurants that you have visited in the past.
\end{itemize}

As a user player, you are expected to:
\begin{itemize}
\itemsep0em
    \item Answer the questions the assistant asks about your preference.
    \item Reject incorrect restaurant recommendations.
    \item Ask questions about the recommended restaurant to justify why you accept or reject the recommendation.
    \item If needed, use the information in your \textit{visited restaurant} to help inform the assistant about your preference.
    \item The frequency of characteristics (or slots) shared by multiple restaurants are indicated in (...), \eg ``(3) parking lot'' means this user has been to 3 restaurants with parking lots.
    \item When you use information from your \textit{visited restaurant}s in one of your responses, make sure to click the  ``Use Fact'' button.
\end{itemize}

\begin{figure}[t]
\centering    
\includegraphics[width=3.0in]{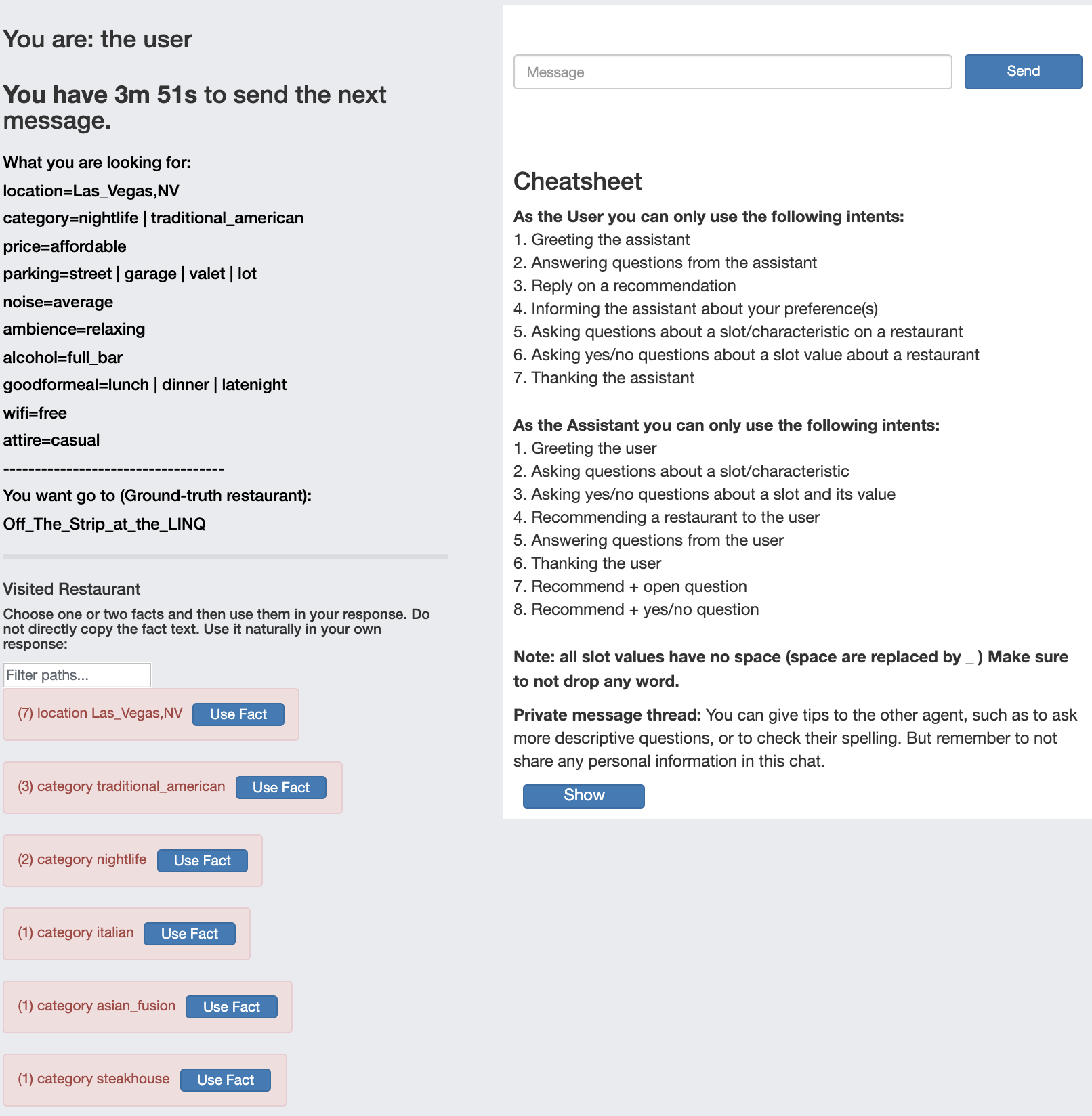}
    \caption{Screenshot of transcription UI for User.}
\label{fig:user}
\vspace{-3mm}
\end{figure}

    

\begin{figure}[t]
\centering    
\includegraphics[width=3.0in]{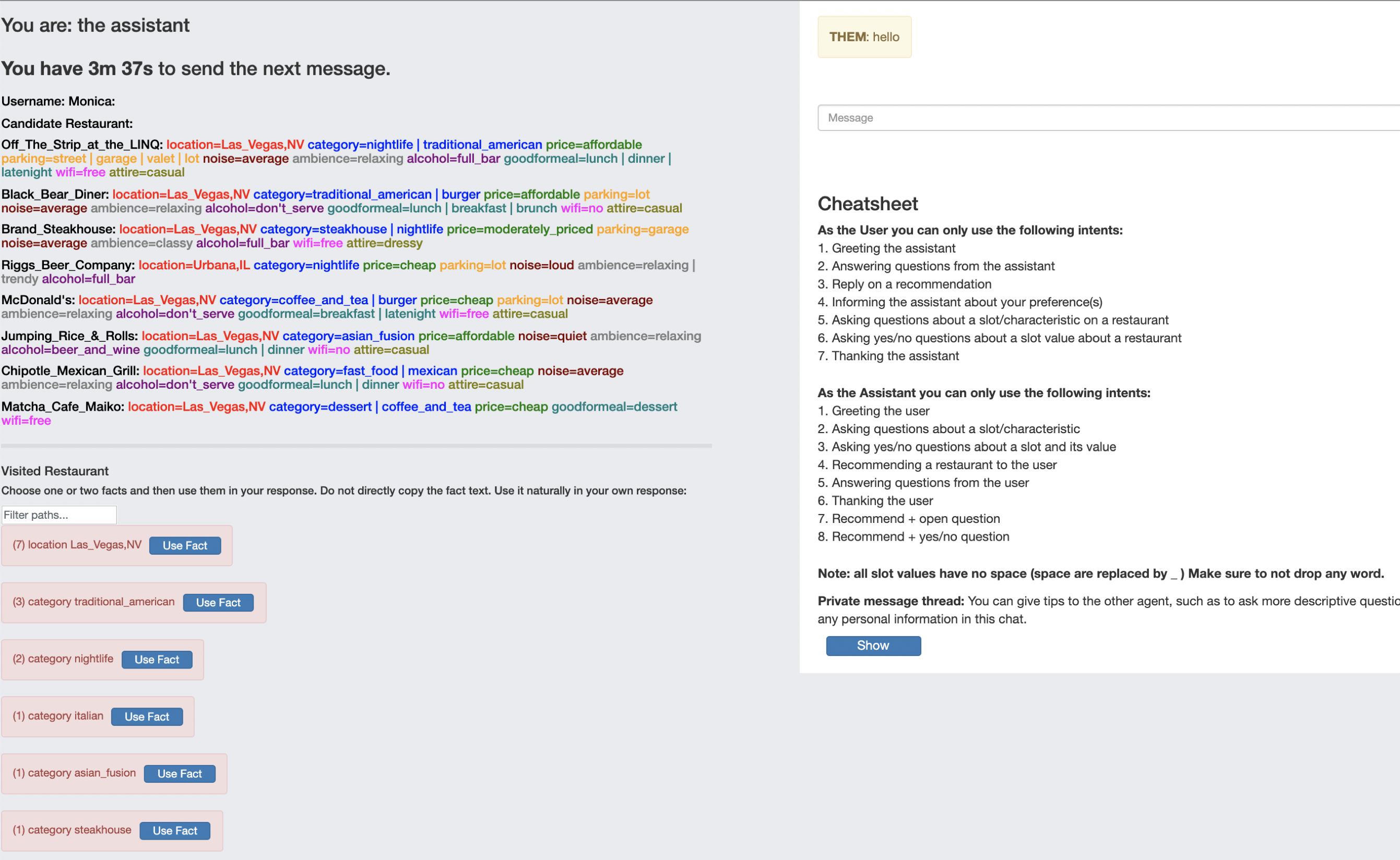}
    \caption{Screenshot of transcription UI for Assistant.}
\label{fig:assistant}
\vspace{-3mm}
\end{figure}

\noindent \textbf{Assistant}: You will interact with the User, to give the desired recommendation (ground-truth restaurant) to the user. You will be provided with the following information:
\begin{itemize}
\itemsep0em
    \item Name of the user.
    \item A list of candidate restaurants, and their characteristics (slots). One of the restaurants in this list is the desired or \textit{ground-truth restaurant} the user is looking for.
    \item Optionally, the characteristics (slots) and values of the restaurants the user has visited (\textit{visited restaurants}). (See the definitions of slots below). The frequency of slots shared by multiple restaurants are indicated in (...), \eg ``(3) parking lot'' means this user has been to 3 restaurants with parking lots.
    The visited restaurants' section may or may not be given to you. If it is given, your goal is to utilize (by clicking ``Use Fact'') the information from \textit{visited restaurant}s as much as possible to provide the desired recommendation to the user. 
\end{itemize}

To make an efficient recommendation, you are expected to: 
\begin{itemize}
\itemsep0em
    \item Ask the user questions about their restaurant preference.
    \item If the visited restaurants are available, \textit{investigate} their slots and values to reduce the number of questions you may need. 
    \item Recommend restaurants to the user based on your knowledge of their preference, their visited restaurants, the information of the candidate restaurants, and from the answers the user gives to your questions.
    \item Intelligently apply the information the user gives to you to guide your conversation.
    \item Recommend the desired restaurant.
\end{itemize}

\subsubsection{Instructions}
This section describes the details of transcription. In general, transcribers are required to follow pre-defined \textit{dialog acts}, \textit{slots} and \textit{values}, but free to make up utterances based on these pre-defined metadata.

\noindent \textbf{Dialog Acts} are the intents of one utterance from a player. 
Note that the user and assistant have their own set of dialog acts, as shown in Table \ref{tbl:dialog_act}. You can only use these pre-defined dialog acts in your utterance.

\noindent \textbf{Slots} refer to 10 pre-defined characteristics of restaurants. 

\noindent \textbf{Values}: one slot is further associated with multiple \textit{values}, such as a slot \textit{Parking} can take value \textit{street}.
Note that a slot can take multiple values at the same time. In the UI, these values are separated by ``$\vert$''. For example, \textit{Parking = street $\vert$ garage} means that a restaurant has both street and garage parking.
DO NOT include ``$\vert$'' in your responses, instead, use one or multiple values naturally in the utterance. \eg ``I prefer street or garage parking.''
You do not have to write out all the values of a slot in one utterance. For example, the \textit{category} slot usually has many values and you do not need to list them all in your utterance.

You will need to write the values exactly as you see them in the UI, including the underscores ``\_'' and commas ``,'' and excluding ``$\vert$''. 
For example, type ``Bonfyre\_American\_Grille'' but not ``Bonfyre American Grille''.
The full lists of values and their slots are at the ends of guidelines. 

\noindent \textbf{Items and their Names}: Each item (restaurant) has an item name and has multiple values and their associated slots. An item is typically associated with a \textit{recommendation} act from the assistant side. 
When recommending a restaurant (item), you are expected to mention the restaurant name (\textit{item name}), which follows the same rule as writing a value in an utterance.

\subsubsection{Important Notes}
During transcribing, it is important to keep these things in mind:
\begin{itemize}
\itemsep0em
    \item A dialog can end with either a user or an assistant response.
    \item The person who plays the user, however, will be the one to terminate the session by pressing the button ``Dialog is done!''
    \item The user should NEVER give all of their preference to the assistant in a single utterance.
    \item The user should NEVER give the \textit{ground-truth restaurant} to the assistant.
    \item When you use content from the \textit{visited restaurant}s in your response, make sure to click the corresponding ``Use Fact'' buttons before sending your response. The click will be recorded.
    \item If the user player has sent more than 10 responses (20 including the responses from the assistant),  it is up to the user player to decide whether to stop the current dialog or to continue.
\end{itemize}

The following actions should be avoided.
\begin{itemize}
\itemsep0em
    \item Do not engage in the transcribed dialog with the other person about the transcription task itself and do not go off-topic. 
    \item Do not share any of your personal information. Always be ``in your character'', i.e., speak as the \textit{user} or the \textit{assistant}.
    \item NO INDECENCY / DISRESPECT / HARASSMENT. Keep your messages decent and respectful towards the other person. Any violations will result in a ban on further tasks.
    \item Do not directly copy any of the utterances from this guideline or UI.
    \item Do not repeat /template your answer, that is to say, do not create one set of responses ahead and then make small changes to them over and over. Please always generate unique and new responses.
\end{itemize}

\subsubsection{Feedback}
After the transcription of one dialog is over, both sides need to give feedback about the transcribed dialog, including:
\begin{enumerate}
\itemsep0em
    \item Rate the dialog (1-5) based on the smoothness and coherence of the whole dialog and whether it closely follows this guideline.
    \item Rate the other side (1-5): whether the other side closely follows this guideline.
    \item (Optional) feedback about this transcription task.
\end{enumerate}

\newpage
\subsection{Annotation Guideline}
\label{sec:transcrib}
In this task, you will get a transcribed dialog between a \textit{user} and an \textit{assistant}, in which the assistant helps the user find the desired restaurant to go to. You will annotate the utterances with \textit{dialog acts}, \textit{slots}, \textit{values}, \textit{item names} and \textit{sentiment} on values or item names. For your reference, the transcription guideline is detailed in Section A. This annotation task will be further supported by a QA process before and during the annotation to resolve hard cases.

\subsubsection{Task}
In this annotation task, you are required to label the following data: 
\begin{enumerate}
\itemsep0em
    \item dialog quality: \textit{good} or \textit{bad} about the whole dialog.
    \item dialog acts (or utterance-level intents), as defined in Table \ref{tbl:dialog_act}.
    \item Label spans of values (or item name) from each utterance and their corresponding slots (or item).
    \item Utterance-level sentiment of each utterance, and optionally span-level sentiment towards a value (or item name) if it is different from the utterance-level sentiment.
\end{enumerate}

You first need to read through the dialog once and label the overall dialog quality, and if it is \textit{good}, label dialog acts.
Then you need to read through the utterances again and label spans of values (or item name), their corresponding slots (item), and sentiment.

\subsubsection{Dialog Quality}
For the entire dialog, you will need to label the dialog quality as either \textit{good} or \textit{bad}. This step is to further ensure the quality of the transcribed dialog. If the dialog quality is labeled as \textit{bad}, you can skip annotating the current dialog further.

\subsubsection{Dialog Acts}
Each utterance must have at least one dialog act. 
The dialog acts are pre-defined in Table \ref{tbl:dialog_act}.
Note that there are different sets of dialog acts for the roles of \textit{user} and \textit{assistant}.
If you believe one utterance is associated with multiple dialog acts, you need to label all of them. We summarize a few important tips for user and assistant separately as following.

\noindent \textbf{Dialog Acts for User}: There are a few key differences among \textit{reply}, \textit{answer}, \textit{inform}, \textit{open} and \textit{yes/no question}.

\begin{itemize}
\itemsep0em
    \item \textit{reply}, \textit{open} and \textit{yes/no question} are always related to a (previously) recommended restaurant (from the assistant). The item name (restaurant name) may or may not show up in the to-be-labeled utterance.
    \item \textit{answer} and \textit{inform} are always related to a value. Note that the value (span) may not show up in an answer (\eg  ``Yes, I like that location.'')
    \item \textit{open question} DOES NOT have a value show up in the utterance but only the explicit or implicit slot (\eg ``what type of food do they serve ?'' [\textit{category}])
    , whereas \textit{yes/no question} must have a value show up (\eg ``do they serve \textit{Italian} food ?'').
    \item \textit{inform}, \textit{open} and \textit{yes/no question} indicates a user actively providing information, while \textit{reply} and \textit{answer} indicate a user passively giving information.
\end{itemize}

\noindent \textbf{Dialog Acts for Assistant}: The key differences among \textit{recommendation}, \textit{open question}, \textit{yes/no question} and \textit{answer} are as following:
\begin{itemize}
\itemsep0em
    \item \textit{recommendation} and \textit{answer} are always related to a restaurant (item). A \textit{recommendation} act may have additional values show up, besides the restaurant name.
    The restaurant typically may not show up in \textit{answer} (\eg ``it serves italian food.'')
    \item \textit{open} and \textit{yes/no question} are always only about slots. But the slot itself may not show up in the utterance directly (\eg ``what kinds of food do they serve ?'').
    \item \textit{yes/no question} always has a value show up: ``do you like \textit{italian} food ?''
\end{itemize}

Note that you always need to annotate the true intent of having an utterance, not the surface form of an utterance. For example, a \textit{recommendation} can have a surface form that looks like a question (e.g., ``how about burger\_king ?'' 
and ``why not try burger\_king ?'').

\subsubsection{Spans of Values, their Slots, Items, and Sentiment}
We expect you to label spans of words that are values (of slots) in the utterance (or item names), all possible values that you can label are listed at the end of this guideline \footnote{We omit the list in this appendix for brevity.}. You are also required to label slots when the values are not shown in an utterance (e.g., an \textit{open question}) by just labeling the slots on utterance-level (similar to label a dialog act). Finally, label utterance-level sentiment (one of \textit{positive}, \textit{negative} and \textit{neutral}) and span-level sentiment (if it differs from utterance-level sentiment).

We expect you to perform the following steps (after you finish labeling dialog acts): 
\begin{enumerate}
\itemsep0em
    \item label spans of words as values (or item names).
    \item select the corresponding slot (or item).
    \item label utterance-level slot (\textit{open question}).
    \item label utterance-level sentiment and check and label span-level sentiment.
\end{enumerate}

\subsubsection{Important Notes}
\begin{itemize}
\itemsep0em
    \item dialog utterances may have typos (e.g., extra spaces, cases), correct and label the spans to the best of your ability, even if errors are present. 
    \item Do not label spans about slots (e.g., \textit{location}, \textit{category}, \textit{price}, etc.) itself, such as words ``where'', ``located at'', ``kinds of'', ``price range'', ``parking'' etc. 
    Labeled spans should only be about pre-defined values or item names (restaurant names).
    \item Utterances from the \textit{assistant} side DO NOT have sentiments.
    \item Only utterances from a user fall in these dialog acts have sentiment: \textit{answer}, \textit{inform}, and \textit{reply}.
    
\end{itemize}

\subsection{Datasets}

\subsubsection{Data Cleaning}
After annotation, the data will go through a data cleaning process via scripts to fix typos and illegal combinations of dialog acts, items, slots, and values. The cleaned data will be integrated with scenarios of each transcription task to form the final datasets.

\subsubsection{Statistics}

\begin{table*}[t]
    \centering
    \scalebox{0.8}{
        \begin{tabular}{c|l|l}
        \hline
        \textbf{Role} & \textbf{Utterance} & \textbf{Acts}\\
        \hline
        User & hello. & Greeting \\
        Assistant & Hello Will! Are you still living in the $\textbf{Phoenix,AZ}_\text{location}$ area. ? & YNQ \\
        User & Yes. [\textit{pos\_on}] & ANS\\
        Assistant & Ok great! Do you want a $\textbf{full bar}_\text{alcohol}$ with your meal? & YNQ\\
        User & No just $\textbf{beer and wine}_\text{alcohol}[\textit{pos\_on}]$ are fine. & ANS\\
        Assistant & My system shows $\textbf{The Nash}_\text{item}$ restaurant. They also offer $\textbf{free}_\text{wifi}$ wifi. & REC\\
        User & Is the food $\textbf{cheap}_\text{price}[\textit{pos\_on}]$ over there? I'm tight on budget. & YNQ, inform \\
        Assistant & It is on the $\textbf{cheap}_\text{price}$ side of the restaurants. & ANS\\
        User & Great. I'll try them out. Thanks. & Reply \& Thanks\\
        Assistant & Thank you and enjoy your meal & Thanks \\
        \hline        
        \end{tabular}
    }
    \vspace{-2pt}
    \caption{An example dialog in MGConvRex with slots and sentiment polarities annotated.} 
    \vspace{-10pt}
\label{tbl:example_dialog}
\end{table*}

\begin{figure}[t]
\centering    
\includegraphics[width=3.2in]{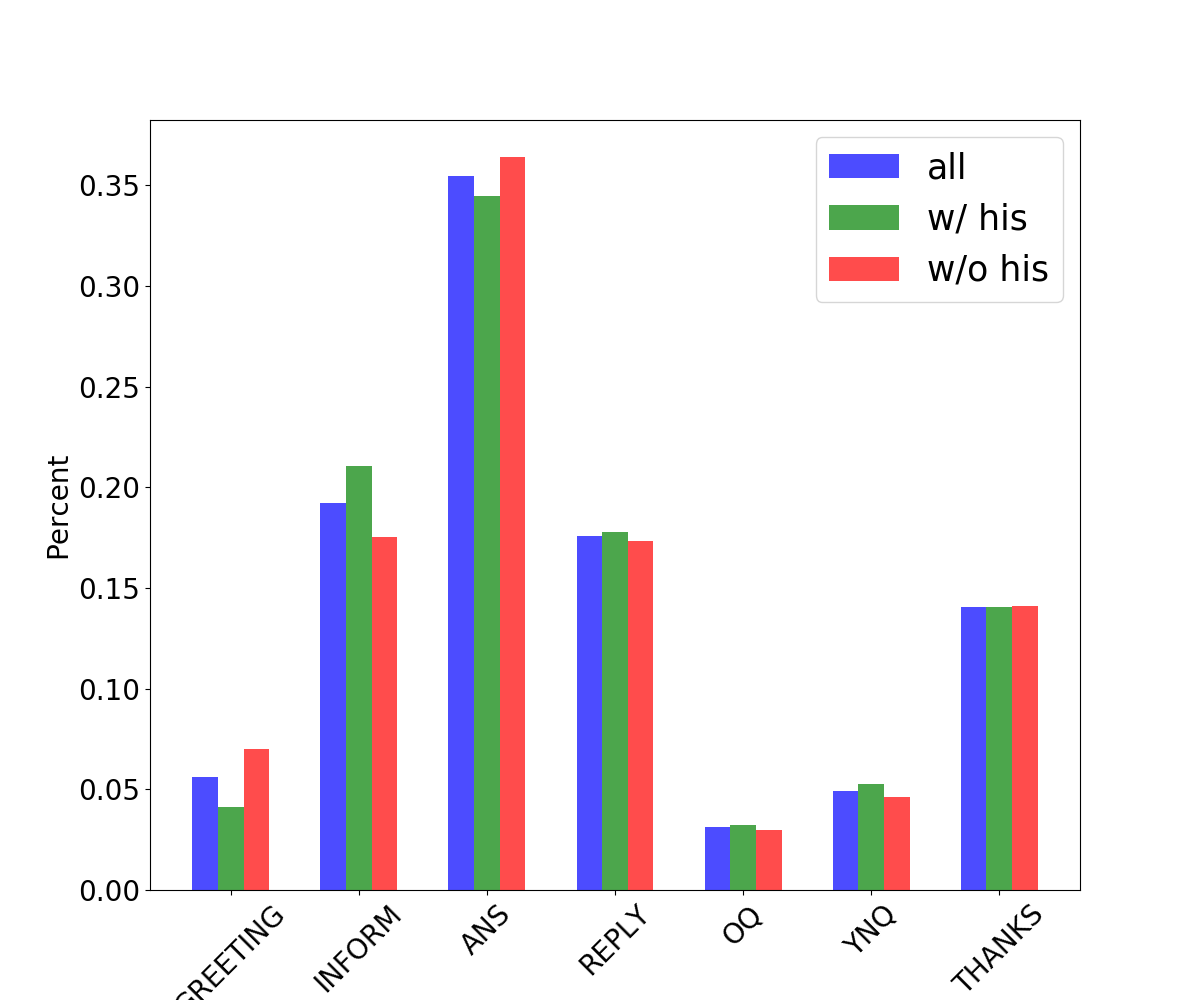}
    \caption{Distribution of user dialog acts}
\label{fig:user_act}
\vspace{-3mm}
\end{figure}

\begin{figure}[t]
\centering    
\includegraphics[width=3.2in]{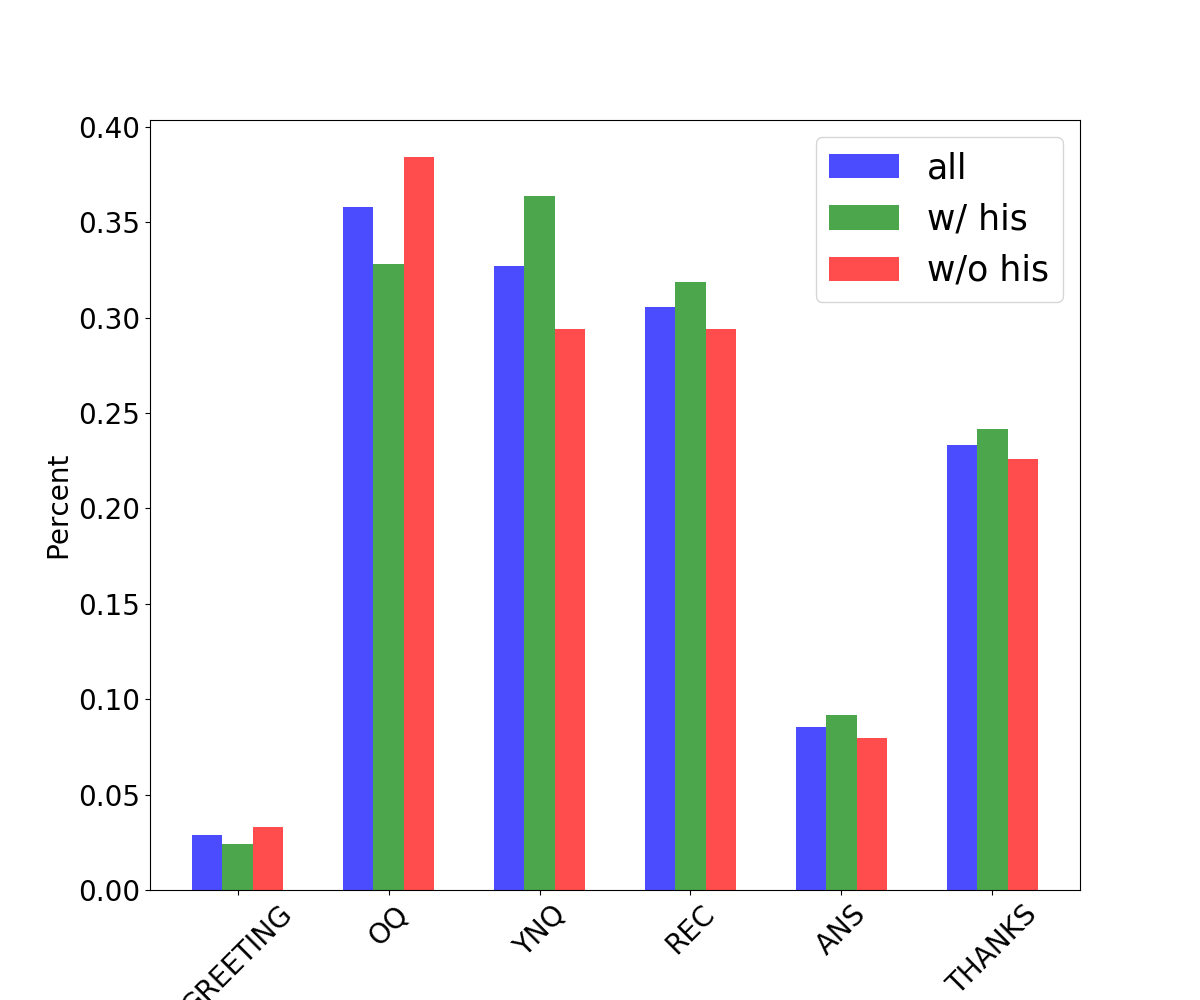}
    \caption{Distribution of agent dialog acts}
\label{fig:agent_act}
\vspace{-3mm}
\end{figure}

\begin{figure}[b]
\centering    
\includegraphics[width=3.2in]{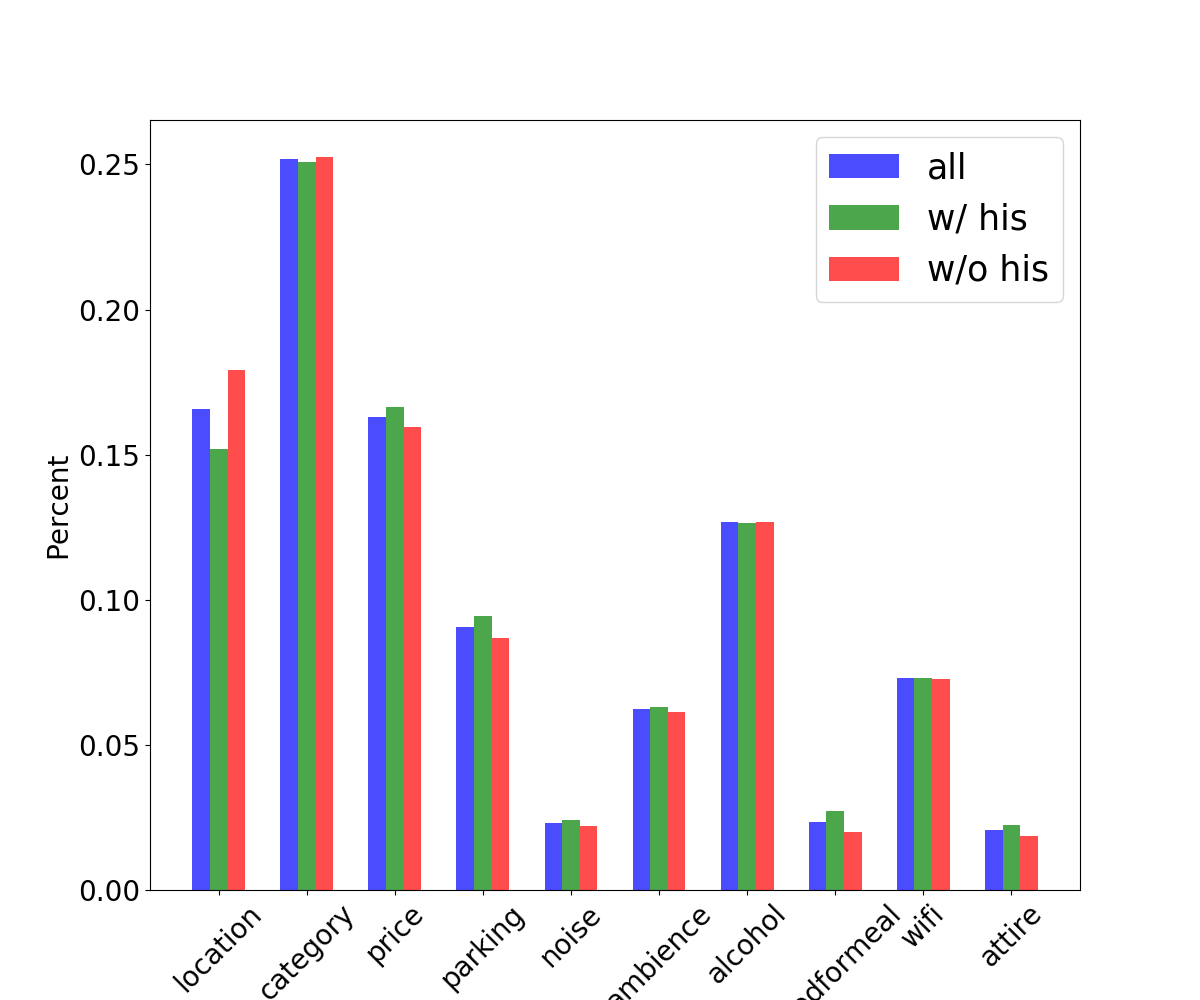}
    \caption{Distribution of slots asked in Open and Yes/no questions from the agent}
\label{fig:agent_slot}
\vspace{-3mm}
\end{figure}

Besides the statistics of MGConvRex in Table \ref{tbl:dataset}, we further study the distributions of dialog acts and slots from the assistant side to learn more about the preferred behavior of crowd workers.
From Figure \ref{fig:user_act}, we can see that a user worker mainly uses the \textit{Answer} act to the agent.
More importantly, the user player is very active and likes to use the \textit{Inform} act without being asked a question. 
User players can sometimes even more cooperative by examining the user memory and inform more salient preference, as indicated by more \textit{inform} in scenarios with history.
From Figure \ref{fig:agent_act}, we can see that an agent worker use both \textit{Open question} and \textit{Yes/no questions} to collect preference. Yes/no questions are more frequent in scenarios with history to confirm users' preferences.
Figure \ref{fig:agent_slot} shows the distribution of slots for \textit{Open and Yes/no question}s asked by the agent player. \textit{category, location, and price} are their mainly used slots for collecting user preference and distinguish different candidate items in $\mathcal{C}$.
We further demonstrate one example dialog is shown in Table \ref{tbl:example_dialog}.

\subsection{User Simulator}
Our online evaluation is conducted against user simulators under a simulation environment in our developed framework.
Here we brief one simulator as in Algorithm \ref{alg:simulator}. 
Note that although our pre-defined dialog acts for user can be either passive or active (as in Table \ref{tbl:dialog_act}), we mostly focus on a passive user (less likely to use \textit{Inform}, \textit{Open / Yes/no question}, \eg 0.2 chance to make an \textit{inform} act) because an active user can vary domain-by-domain and hard to implement. In the implementation, the user mostly follows the dialog acts from the agent and provide information accordingly in a case-by-case fashion.

\begin{algorithm}[t]
\LinesNumbered
\DontPrintSemicolon
\caption{Algorithm for User Simulator}
\label{alg:simulator}
\SetKwInOut{Input}{Input} 
\SetKwInOut{Output}{Output} 
\SetKwProg{Def}{def}{:}{}
\Input{$a$, $e_i$, $e_s$, $e_v$ from the agent; $P$ and $\mathcal{T}$ from scenario}
\Output{$a'$, $e'_i$, $e'_s$, $e'_v$, $o'$ from the user}

\BlankLine

\Def{$\text{RandomGreetInform}()$}{
    \If{$\text{random} < 0.8$}{
        $a' \gets \textit{GREETING}$ \;
    }
    \Else{
        $a' \gets \textit{INFORM}$ \;
        $e'_v, o' \gets \text{RandomValue}(), \textit{pos\_on}$\;
    }
    \Return $a', e'_v, o'$ \;
}
\BlankLine
$a', e'_i, e'_s, e'_v, o' \gets \text{Nones}$ \;
\Switch{$a$}{
    \Case{$\text{INIT}$ \tcp*{first turn}}{
        $a', e'_v, o' \gets \text{RandomGreetInform}()$\;
    }
    \Case{$\text{REC}$}{
        \If{$e_i \in \mathcal{T}$}{
            $a', o' \gets \textit{REPLY}, \textit{pos\_on}$ \;
            $\text{SetDialogSuccess}()$ \;
        }
        \Else{
            $a', e'_s \gets \textit{OQ}, \text{RandomSlot}()$ \;
        }
    }
    \Case{$\text{OQ or YNQ}$}{
        $e'_v, o' \gets \text{FindValueOpinion}(P)$ \;    
    }
    \Case{$\text{ANS}$}{
        \If{$P \text{ has } e_v$}{
            $a', o' \gets \textit{INFORM}, \textit{pos\_on}$ \;
        }
        \Else{
            $a', o' \gets \textit{INFORM}, \textit{neg\_on}$ \;
        }
    }
    \Case{$\text{THANKS}$}{
        \If{$\text{IsDialogSuccess}() $}{
            $a', o' \gets \textit{THANKS}, \textit{pos\_on}$ \;
        }
        \Else{
            $a', e'_v, o' \gets \text{RandomGreetInform}()$ \;
        }   
    }
}

\Return $a', e'_i, e'_s, e'_v, o'$ \;

\end{algorithm}

\end{document}